\documentclass[final,5p,times,twocolumn]{elsarticle}

\usepackage{hyperref}
\usepackage{tikz}
\usetikzlibrary{mindmap,trees}
\usetikzlibrary{arrows,positioning,calc,decorations.markings}
\usepackage{subcaption}
\usepackage{balance}

\makeatletter
\def\ps@pprintTitle{%
	\let\@oddhead\@empty
	\let\@evenhead\@empty
	\def\@oddfoot{\footnotesize\itshape Published in Robotics and Autonomous Systems\hfill \today}%
	\let\@evenfoot\@oddfoot}
\makeatother

\journal{Robotics and Autonomous Systems}

\begin{document}

\begin{frontmatter}

\title{A Survey on Human-aware Robot Navigation}

\author[1]{Ronja Möller}
\author[1]{Antonino Furnari}
\author[1]{Sebastiano Battiato}
\author[2]{Aki Härmä}
\author[1,3]{Giovanni Maria Farinella\corref{cor1} }
\cortext[cor1]{Corresponding author:}
\ead{gfarinella@dmi.unict.it}

\address[1]{Department of Mathematics and Informatics, University of Catania, V. Andrea Doria, Catania 95125, Italy}
\address[2]{Philips Research, High Tech Campus 34, 5656 Eindhoven, Netherlands}
\address[3]{Cognitive Robotics and Social Sensing Laboratory, ICAR-CNR, Palermo, Italy}

\begin{abstract}
Intelligent systems are increasingly part of our everyday lives and have been integrated seamlessly to the point where it is difficult to imagine a world without them. Physical manifestations of those systems on the other hand, in the form of embodied agents or robots, have so far been used only for specific applications and are often limited to functional roles (e.g. in the industry, entertainment and military fields). Given the current growth and innovation in the research communities concerned with the topics of robot navigation, human-robot-interaction and human activity recognition, it seems like this might soon change. Robots are increasingly easy to obtain and use and the acceptance of them in general is growing. However, the design of a socially compliant robot that can function as a companion needs to take various areas of research into account. This paper is concerned with the navigation aspect of a socially-compliant robot and provides a survey of existing solutions for the relevant areas of research as well as an outlook on possible future directions.
\end{abstract}


\end{frontmatter}

\section{Introduction and Motivations}

Robots have come a long way from initial attempts to create artificial humanoids that could walk and or talk\footnote{\url{http://cyberneticzoo.com/robots/1928-eric-robot-capt-}\\\url{richards-english/}, \\\url{https://spectrum.ieee.org/tech-history/dawn-of-} \url{electronics/elektro-the-motoman-had-the-biggest-brain-} \url{at-the-1939-worlds-fair}. } to more functional designs that are truly autonomous, such as the work of Boston Dynamics\footnote{\url{https://www.bostondynamics.com/robots}}, a spin-off from the Massachusetts Institute of Technology on one side, and near-lifelike copies of humans such as ``Geminoid"\footnote{\url{http://www.geminoid.jp/projects/kibans/resources.html}}, built by Hiroshi Ishiguro Laboratories, on the other side. While early robots served as outlooks for future technology or performed very simple tasks designed to impress the audience, more recent research has produced an increasingly diverse number of robots that can perform more and more complex tasks independently of user input. The inclusion of robots in industrial settings, beginning with a digitally programmable robot that lifted and stacked metal pieces by George Devol in 1954\footnote{\url{https://www.robotics.org/joseph-engelberger/unimate.cfm}}, has fundamentally changed virtually every industry (see, e.g., \cite{ford_rise_2015} for an excellent thought-provoking summary). Robots are now widely used in manufacturing, assembly, packing, transport, search and rescue, healthcare, surgery and laboratory research to name just a few. \\
The usage of robots in social contexts, on the other hand, is still in an earlier stage. A few companies have successfully marketed robots as pets, with the Tekno (a small robotic dog) being one of the more popular, and recently companion robots like Pepper\footnote{\url{https://www.softbankrobotics.com/emea/en/pepper}}, Jibo\footnote{\url{https://jibo.com/}} and Loomo\footnote{\url{https://de-de.segway.com/products/segway-loomo-robot}} have been presented. Even with these successes, none of them have reached the popularity of intelligent non-embodied systems such as, for example, the Amazon Alexa.\\ 
In addition to being useful and comfortable to be around, a social robot naturally has to be affordable for the demographic it is meant for. While institutions such as hospitals or hotels can afford more sophisticated and expensive robots, the ``standard" user - and researcher - will be constrained by a certain budget. As a result, the most commonly used sensors in research tend to be affordable options such as simple distance or proximity sensors (ultrasound, infrared and sometimes LiDAR - Light Detection and Ranging), odometry sensors, GPS (Global Positioning System) and images/video. Image sensors in particular are very popular across all kinds of applications and price ranges. This survey will focus on image-based research.\\
As soon as an embodied agent starts moving in the user's space and interacting with them, a variety of social rules has to be obeyed, as discussed in the book by Knapp et al. on  Non-Verbal Communication in Human Interaction \cite{knapp_non-verbal_2010}. The robot's behaviour has to be carefully designed to be human-like enough to inspire trust and confidence without erring into the uncanny valley territory \cite{mori_uncanny_2012}, risking user rejection. The way a robot moves is a particularly important aspect of user acceptance, with a variety of studies evaluating how the speed of movement, observed distances and cultural biases influences the user's perception of a robot \cite{pacchierotti_evaluation_2006}, \cite{henkel_evaluation_2014}, \cite{kim_how_2014}. \\
Given the interdisciplinary nature of this task, solutions have to fuse the insights of several different areas and deal with the advantages and shortcomings of each. Robot Navigation, for example, is a well-established field of research with a variety of available metrics \cite{anderson_evaluation_2018}, benchmarks and datasets \cite{chang_matterport3d_2017} \cite{ammirato_dataset_2017} \cite{kolve_ai2-thor_2017} \cite{savva_habitat_2019} . Human-robot-interaction on the other hand is more subjective and evaluated under different criteria \cite{young_evaluating_2011}, making joint datasets or evaluations difficult. The necessary inclusion of human activity recognition (and potentially emotion recognition) - to enable the robot to move in reaction to the action its user is currently performing (and its mood) - further compounds the issue. Even with these difficulties, researchers are motivated to tackle this fascinating problem.\\
Artificial Intelligence on an embodied social companion has the potential to improve the quality of life, equalize disparities for disabled users and, as argued by Topol in his book on deep learning in medicine \cite{topol_deep_2019}, make healthcare more ``human" by solving menial tasks so the caregivers have more time for their patients. Social robots are also expected to have an important role in motivation and patient engagement, in the self-management of a chronic disease and adherence to therapies and medication\footnote{\url{https://www.mobihealthnews.com/content/catalia-health-} \url{ gets-25m-improve-medication-adherence-social-robot}}. While these ideas seem lofty and far-fetched, every embodied agent trying to accomplish those goals will have to move in a way that is socially compliant, hence solving this task will always be the first step.
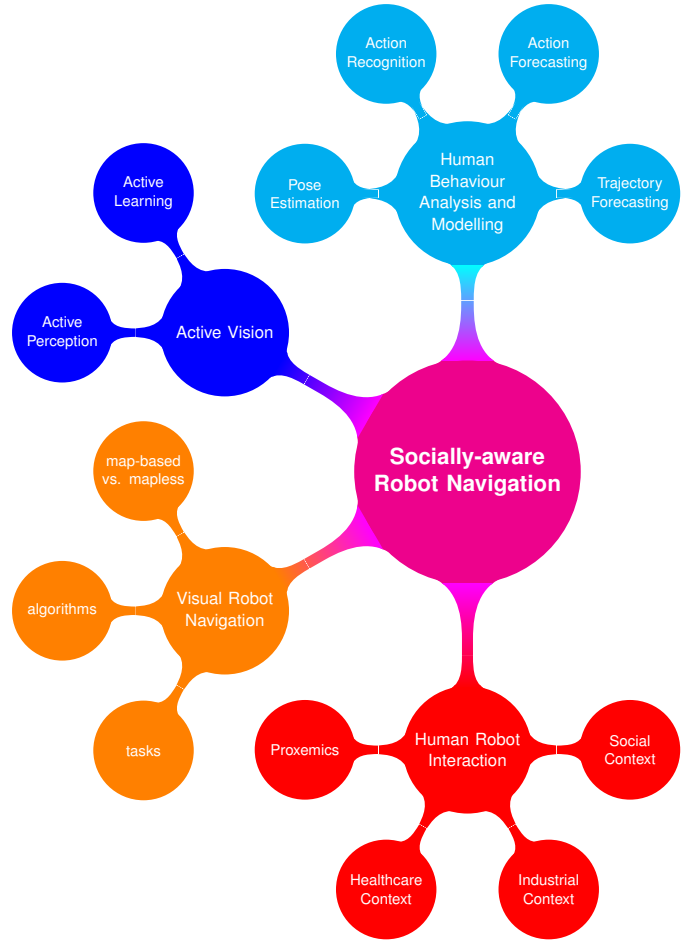
\begin{figure}[ht]
\resizebox{\columnwidth}{!}{%
\begin{tikzpicture} [ every annotation/.style = {draw, fill = white, font = \Large}]
\iffalse
  \path[mindmap,concept color=magenta,text=white, font=\large\bfseries]
    node[concept] {\textsf{Socially-aware Robot Navigation}}
    [clockwise from=90]
    child[concept color=blue] { node[concept] {\textsf{Active Vision}}
     [clockwise from=120]
      child { node[concept] {\textsf{Active Perception}} }
      child { node[concept] {\textsf{Active Learning}} }}
    child[concept color=orange] {node[concept] {\textsf{Visual Robot Navigation}}
      [clockwise from=120]
      child { node[concept] {\textsf{tasks}} }
      child { node[concept] {\textsf{algorithms}} }
      child { node[concept] {\textsf{map-based vs. mapless}} }}
    child[concept color=red] {node[concept] {\textsf{Human Robot Interaction}}
      [clockwise from=60]
      child { node[concept] {\textsf{Social Context}} }
      child { node[concept] {\textsf{Industrial Context}} }
      child { node[concept] {\textsf{Healthcare Context}} }}
    child[concept color=cyan] {node[concept] {\textsf{Human Behaviour Analysis and Modelling}}
      [clockwise from=0]
      child { node[concept] {\textsf{Pose Estimation}} }
      child { node[concept] {\textsf{Action Recognition}} }
      child { node[concept] {\textsf{Action Forecasting}} }
      child { node[concept] {\textsf{Trajectory Forecasting}} }}
    child[concept color=yellow!90!black] {node[concept] {\textsf{Human-aware Motion Planning}}
    [clockwise from=-120]
      child { node[concept] {\textsf{Proxemics}} }
      child { node[concept] {\textsf{Human Following}} }
      child { node[concept] {\textsf{Moving among Humans}} }};
\fi
\path[mindmap,concept color=magenta,text=white, font=\large\bfseries]
    node[concept] {\textsf{Socially-aware Robot Navigation}}
    [clockwise from=-90]
    child[concept color=red] {node[concept] {\textsf{Human Robot Interaction}}
      [clockwise from=0]
      child { node[concept] {\textsf{Social Context}} }
      child { node[concept] {\textsf{Industrial Context}} }
      child { node[concept] {\textsf{Healthcare Context}} }
      child { node[concept] {\textsf{Proxemics}} }}
    child[concept color=orange] {node[concept] {\textsf{Visual Robot Navigation}}
      [clockwise from=240]
      child { node[concept] {\textsf{tasks}} }
      child { node[concept] {\textsf{algorithms}} }
      child { node[concept] {\textsf{map-based vs. mapless}} }}
    child[concept color=blue] { node[concept] {\textsf{Active Vision}}
     [clockwise from=180]
      child { node[concept] {\textsf{Active Perception}} }
      child { node[concept] {\textsf{Active Learning}} }}
    child[concept color=cyan] {node[concept] {\textsf{Human Behaviour Analysis and Modelling}}
      [clockwise from=180]
      child { node[concept] {\textsf{Pose Estimation}} }
      child { node[concept] {\textsf{Action Recognition}} }
      child { node[concept] {\textsf{Action Forecasting}} }
      child { node[concept] {\textsf{Trajectory Forecasting}} }}
   ;
\end{tikzpicture}
}
\caption{ Socially-aware robot interaction is a very complex topic and interdisciplinary by design. It covers robot navigation tasks, social and cultural rules as well as Human Behaviour Analysis. }
\label{overview}
\end{figure}
This survey will consider active vision, robot navigation, human-robot interaction and human activity recognition as well as their possible overlaps. Several very informative surveys cover robot navigation \cite{gul_comprehensive_2019}, \cite{patle_review_2019}, \cite{hoy_algorithms_2015}, \cite{yang_survey_2016} - some with a focus on visual input \cite{desouza_vision_2002}, \cite{bonin-font_visual_2008}, \cite{hirschmuller_computer_2015} - and the topics of human-robot interaction \cite{goodrich_human-robot_2007}, \cite{yan_survey_2014}, \cite{cherubini_collaborative_2016}, \cite{villani_survey_2018-1} and human action recognition have been surveyed extensively as well \cite{moeslund_survey_2006}, \cite{aggarwal_human_2011}, \cite{herath_going_2017}, \cite{zhang_comprehensive_2019}. We will review the topic of robot navigation with a focus on recent trends and approaches that can be used in settings where humans will be encountered.\\
Previous surveys on socially aware robot navigation focused purely on the social requirements \cite{kruse_human-aware_2013}, such as comfort, naturalness and cultural conventions, or the type of mapping that was used for navigation \cite{charalampous_recent_2017}, such as metric mapping, semantic mapping and social mapping. Our aim with this work is to give a more general overview and highlight the interdisciplinary nature of the topic. \\
The preferred input sensor in most surveyed approaches will be RGB(D) (Red Green Blue (Depth)) cameras, but we will include publications that use different sensors if they have influenced later visual approaches in terms of strategies and evaluations. We will also cover popular concepts of human-robot interaction, concentrating on those relevant to everyday surface-level social ``interaction" such as navigating around or with humans. However, the survey will not focus on actual social interaction such as talking, touching, and collaborating. For human-robot interaction in industrial settings, we would like to suggest~\cite{villani_survey_2018} for a relatively recent overview. For a robot to interact naturally and economically with a human user it is also important to take into account what action is currently performed by the human(s). To this end, we will look into work on activity recognition that uses visual or depth sensors as input. We will not go into depth on alternatives like inertial sensors or RFID (Radio-frequency identification) tags.\\ We believe that in order for a socially-aware robot to behave ``correctly", it is important to combine insights from all four topics - namely active vision, robot navigation, human-robot interaction and human behaviour modeling - this survey that covers all these topics in conjunction, to inspire further research. \\
We will start with an introduction to active vision as the paradigm underlying all research with embodied active agents, then we will cover the topic of visual robot navigation and its most popular strategies, measures and current state of the art. In the later sections, the focus will be on ``the human factor" with introductions to the topics of Human-Robot-Interaction, the analysis and prediction of human movement and human behaviour.  In each of these sections we will also highlight works that have successfully implemented socially aware navigation strategies that include the respective information (Activity Recognition, Trajectory Prediction etc.) The last section will discuss observed issues and research gaps and give outlooks on potential future directions. 
Figure \ref{overview} provides a visual representation of the topics covered by this survey.
 
\section{The Active Vision paradigm}
We will first give a brief introduction to the difference between the active and passive vision paradigms and how they relate to the tasks of classification, segmentation and navigation. The second section will go into detail about specific subsets of the active vision paradigm and how they can be applied to concrete problem settings.
\begin{figure}
\resizebox{\columnwidth}{!}{%
\includegraphics[width=\textwidth]{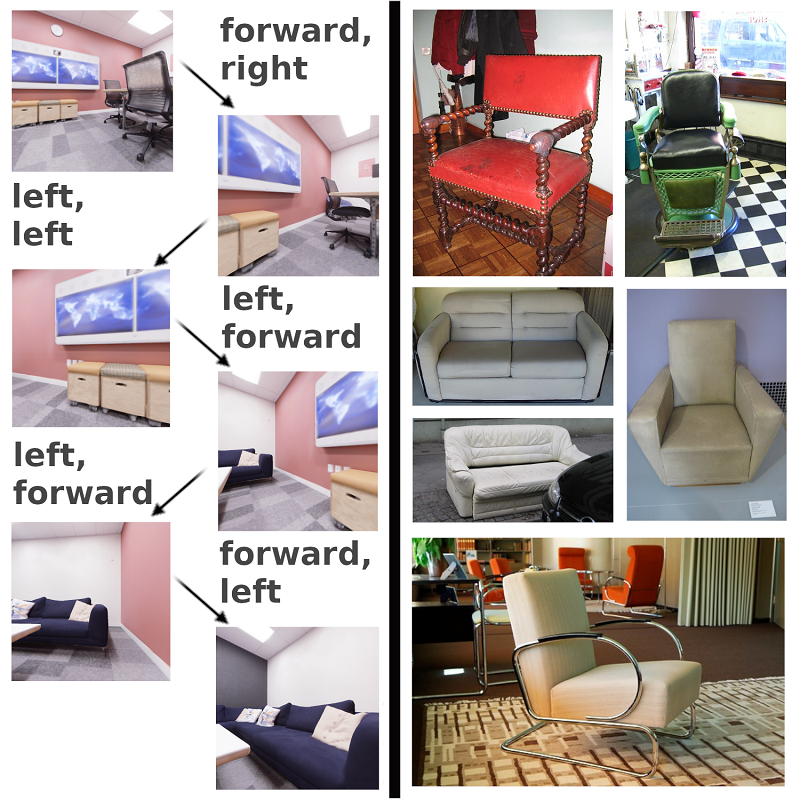}
}
\caption{ Illustration of the difference in input for a an embodied agent (left half, AI Habitat simulator \cite{savva_habitat_2019}) that encounters chairs and sofas in an office space vs. samples of a standard visual learning training set (right half, ImageNet~\cite{krizhevsky_imagenet_2012}) for the object class seating (chairs, sofas).  }
\label{activa_vision_img}
\end{figure}
\subsection{Active vs. Passive Vision}

Simply put, an Active Vision System can influence the input by controlling the camera, while a Passive Vision System cannot. This is a special case of Active Learning, which differs from Passive Learning in that the algorithm actively and intentionally chooses which samples to train on. In Passive Learning and by extension Passive Vision, the input data is chosen only by the researcher and the algorithm simply processes each given sample.\\

Instead of learning the correspondence between a fixed set of inputs and the desired outputs (for example classification or segmentation), an Active Vision System learns advantageous behaviour during training and or testing. In addition, through active learning, an agent can learn underlying dependencies or hierarchies that cannot be captured by learning from independent datapoints. A concrete example would be an agent that learns dependencies of objects and rooms, such as: fridge and microwave are in the kitchen, toilet and shower are in a different room. These dependencies and hierarchies can only be learned through either explicit modeling or active exploration of the sample space.\\
\paragraph{Differences in input data}

The added complexity of an Active Vision System is also motivated by the fact that vision applications on robots, especially robots capable of movements, will always have to take into account that the input data differs a lot from standard classification datasets. While curated datasets such as ImageNet \cite{krizhevsky_imagenet_2012} and MS COCO \cite{lin_microsoft_2014} have undoubtedly greatly pushed the state of the art for image classification and segmentation, they contain samples that are well curated and mainly acquired from a third person perspective. This assumption does not hold for what a robot will see ``in the wild" and an argument can be made that this way of learning also does not correspond to how humans learn. Indeed, humans learn not with large numbers of perfectly centered and lit samples but instead with fewer samples that are looked at from different directions. Moreover, humans are embodied agents, which allows them to continuously interact with the world they learn from. An example for this difference would be an algorithm that learns about an object by being fed thousands of image examples versus a robot navigating around objects of the specific type, see Figure \ref{activa_vision_img}.  Not only will the input data look different in terms of orientation, occlusion, lighting etc. but it contains the added information about the actions the robot performed to arrive at the new input. In contrast, the samples on the right are all centered, unobstructed, from similar perspectives and independent with no relationship between the samples. \\
Starting from the outlined differences, two ways of handling the problems come to mind. One could either accept that the input data will be incomplete/degraded and learn strategies to cope - for example temporal smoothing, consistency checks and common sense rules. The other approach would be to accept that the issue is intrinsic to having an embodied agent and instead learn behaviour that improves the input quality and therefore yields a better results. In other words, the freedom of movement can be seen as a feature, rather than a problem.  While both options are viable for many problem settings, such as classification, detection and segmentation, the task of object navigation tends be solved with the second one. \\
\paragraph{Examples for Passive Vision}
There are successful works on robot navigation that use a straightforward approach of  matching input data to a desired output: Bojarski et al. \cite{bojarski_end_2016} from NVIDIA proposed an end-to-end solution for self driving cars by mapping raw RGB input from a single front-facing camera on a car to directions. In a similar vein, Giusti et al. \cite{giusti_machine_2015} created their own dataset (with several viewpoints) and used it to map an image to the {\sl correct} direction when navigating along a wilderness trail. A current iteration of this more ``passive" learning approach was proposed by Chang et al. \cite{chang_semantic_2020}, who used the implicit movement labels on real estate Youtube videos to train an agent. \\

\paragraph{Examples for Active Vision}
In contrast, there are several arguments for movement during training and testing. A very early work from the sixties by Held and Hein \cite{held_movement-produced_1963} suggested that kittens, as a standin for intelligent agents who navigate using visual input, can only learn the correspondence between movement and visual input by physically performing this movement, not by merely observing changing environments. In their experiments, a set of kittens who where restricted in an apparatus to be moved passively, displayed worse movement capabilities compared to kittens who were allowed to actively move. Bajcsy et al. \cite{bajcsy_revisiting_2016} make a convincing argument why any research on robot navigation necessarily has to include active perception, to quote their conclusion: ``An agent is an active perceiver if it knows why  it  wishes  to  sense,  and  then  chooses  what  to  perceive,  and  determines  how,  when  and  where  to achieve that perception". An active solution for autonomous automotive navigation has been proposed by Kendall et al. \cite{kendall_learning_2019}, whereas an active approach for robot navigation was presented by Morad et al. \cite{morad_embodied_2020}.\\
The addition of a social aspect is another dimension that is very hard to learn in a passive way: it is challenging to definitively decide on a ``correct" motion in relation to humans since they react differently to the presence of the robot and have their own specific behaviours as well \cite{mead_perceptual_2016}, \cite{pacchierotti_evaluation_2006}, \cite{mavrogiannis_effects_2019}. \\

The active paradigm will be present in most of the embodied robot research in this paper. Although the survey will show how many passive algorithms can be ``re-purposed" to fit this viewpoint, the fundamental difference between active vision and the standard ``passive" approach is still relevant. \\
\subsection{Active Perception and Active Learning}
Active Perception refers to a setting in which the robot/sensor can move during testing to gain more useful information and thereby improve the performance. For a summary of active perception we recommend the original introduction by Aloimonos et al.~\cite{aloimonos_active_1988} from the eighties and the updated review by the same authors~\cite{bajcsy_revisiting_2016}, as well as the more recent work by Chen et al.~\cite{chen_active_2011}.
The applications of this mindset are also not necessarily limited to robotic navigation tasks, but can also extend to exploratory behaviour and image inference or hallucination - see the work of Grauman et al. \cite{ramakrishnan_emergence_2019}, \cite{jayaraman_learning_2018} - and object category detection - see Ammirato et al. \cite{ammirato_dataset_2017} who used a novel dataset to simulate active vision. \\
In contrast to Active Perception, Active Learning \cite{settles_active_2010} in a robotic context is more concerned with exploiting ``movement" (as in, the ability to actively influence the input) during training, enabling the agent to intelligently choose what datapoints to exploit next. Applications include video segmentation, as proposed in Fathi et al. \cite{fathi_combining_2011}, image classification on intelligently chosen data subsets as in Sener et al.  \cite{sener_geometric_2017}, and medical image analysis (see Mahapatra et al's \cite{mahapatra_semi-supervised_2013} work on Crohn disease). Another interesting and very recent work by Chaplot et al.~\cite{chaplot_semantic_2020} combines embodied robot navigation with active vision to finetune a pretrained network for a segmentation task.\\

\section{Visual Robot Navigation}
The navigation of mobile robots has been of great academic interest for over 30 years and the field has continued to change with advancements in hardware and algorithms \cite{desouza_vision_2002}. Depending on the design and purpose of the robot, different strategies for locomotion have been proposed \cite{lobo_sensors_1998}. For a comprehensive summary of the sensors used before 2000, we recommend the surveys by Lobo et al. \cite{lobo_sensors_1998} and Borenstein et al. \cite{borenstein_where_1996}. The main driving industries for autonomous sensing in recent times have been the military and the automotive sector and the latter is very open about extensively using cameras \cite{yurtsever_survey_2020}. This trend combined with the increased demand for complex and ``human-like" sensing for assistive robots and the decreased price of commercial cameras has made visual robot navigation very appealing. \\
Given the historic progression of the field, there are different ways to break down the overall topic and these divisions keep changing with the preferred sensors and applications. In this survey, we will keep the distinction between map-based and mapless navigation from \cite{desouza_vision_2002} and \cite{bonin-font_visual_2008} with an emphasis on mapless navigation since our focus is on navigation in relation to humans in a variety of possibly unstructured or unknown environments. We will not be separating approaches into indoor-vs outdoor navigation like the mentioned surveys as the most socially relevant interactions tend to happen indoors. Instead, we will present some navigation tasks that are currently popular in the academic community and how they relate to human-aware navigation. \\
Although several of the presented problem settings do not contain humans or treat them as static obstacles only, they are convenient to study general navigation abilities, which are fundamental for social robots. Moreover, many strategies from classic navigation have been adapted to include a social dimension while using the same underlying algorithms.\\

\subsection{Map-based vs. Mapless Navigation}
\label{Map-based vs. Mapless Navigation}
When designing a robot and the accompanying algorithms, one of the earliest design decisions to be made is the type of environment the robot should operate in and whether or not it should create an internal representation of it. Benefits of internal representations (in the form of maps) are that the robot can navigate faster to areas it has already seen and therefore operate in a more economic way. An example where this type of behaviour is desired would be an industry robot that only operates in a certain unchanging setting and is always able to compute the shortest path to a goal in the known environment. In less controlled, less structured or non-static environments such as public spaces or homes with moving pedestrians, the creation of a fixed map might be impossible (due to complexity or change). As a result, it might be more useful to instead learn general behaviours without an internally maintained map. An example for this type of behaviour would be a robot that learns to navigate in unknown scenes by minimizing collisions. \\
Earlier work focused on metric or grid-like internal map presentations (see for example the highly influential work by Moravec et al. \cite{moravec_high_1985}, where a sonar is used to generate an occupancy grid). The robot can navigate among non-occupied space on the map. Further research has investigated other map representations. For example, the authors of \cite{kuipers_robot_1991} and~\cite{engelson_error_1992}, who introduced the concept of topological maps consisting of interconnected landmarks in a graph-like structure that encode traversability in the edges.\\
\begin{figure}[t]
\resizebox{\columnwidth}{!}{%
\begin{tikzpicture}
\centering
\draw[black,thick] (1,0) -- (4,0);
\draw[black,thick] (4,1) -- (4,2.5);
\draw[black,thick] (4,3.5) -- (4,4) -- (3,4);
\draw[black,thick] (2,4) -- (1,4) -- (1,0);
\draw[dotted]
    (2.5,3) node {living};
\draw[dotted]
    (2.5,1) node {kitchen};
\draw[black,thick] (1,4) -- (1,5) -- (7,5) -- (7,4);
\draw[dotted]
    (4,4.5) node {hall};
\draw[black,thick] (4,0) -- (7,0) -- (7,2) -- (6.5,2);
\draw[black,thick] (4,2) -- (5.5,2);
\draw[dotted]
    (5.5,1) node {sleep};
\draw[black,thick] (7,2) -- (7,4) -- (4,4);
\draw[dotted]
    (5.5,3) node {bath};

\draw[black,thick] (10,2.5) -- (12,1);
\draw[black,thick] (12,4) -- (12,1);
\draw[black,thick] (12,4) -- (15,1);
\draw[black,thick] (15,1) -- (15,4);
\draw[black,thick] (12,1) -- (15,4);

\fill[white] (10,2.5) circle (0.7 cm);
\draw (10,2.5) circle (0.7 cm);
\draw[dotted]
    (10,2.5) node {hall};

\fill[white] (12,1) circle (0.7 cm);
\draw (12,1) circle (0.7 cm);
\draw[dotted]
    (12,1) node {living};
\fill[white] (12,4) circle (0.7 cm);
\draw (12,4) circle (0.7 cm);
\draw[dotted]
    (12,4) node {kitchen};
\fill[white](15,1) circle (0.7 cm);
\draw (15,1) circle (0.7 cm);
\draw[dotted]
    (15,1) node {sleep};
\fill[white] (15,4) circle (0.7 cm);
\draw (15,4) circle (0.7 cm);
\draw[dotted]
    (15,4) node {bath};

\end{tikzpicture}
}
\caption{An example of an ideal metric map (left) that could be created by a SLAM algorithm vs. the same environment encoded in a topological map (right).}
\label{maps}
\end{figure}
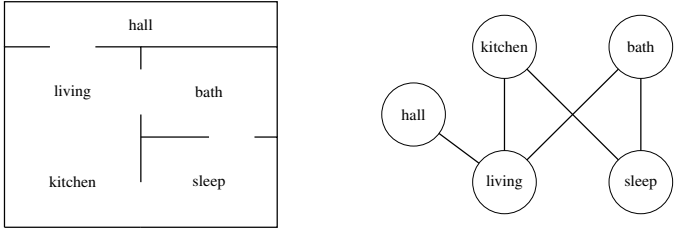
Figure \ref{maps} illustrates different map representations for the same environment.  The edges between the vertices of the topological map denote accessability/traversability. The rooms themselves can be encoded in the topological map in various ways, for example using visual similarity or activities that are performed frequently.\\
Metric maps are intuitive and provide a good basis for metric optimization problems like shortest path computations, but they are difficult to build, very inefficient for larger spaces and their usability degrades the less accurate the robot position is. Topological maps on the other hand can be more efficient, especially for simple environments, they don't require the robot's position to be very exact and they support more high-level planning approaches. The drawbacks of topological maps are ambiguity, their high complexity for articulated or cluttered environments and that they cannot be used for metric path optimization, but only for symbolic approaches. This division does not have to be strict either, as seen in the work by Thrun et al. \cite{thrun_learning_1996}. Although the mentioned works focus on distance sensors and precede vision-based approaches, more current approaches that use cameras still use those paradigms. For a comprehensive overview on map-based navigation that is somewhat sensor-agnostic we recommend the two outstanding reviews by Meyer Filliat~\cite{filliat_map-based_2003} \cite{meyer_map-based_2003} that include camera and non-camera sensors but also discuss approaches independently of the actual sensors.\\
Navigation systems can also rely on SLAM (Simultaneous Localisation and Mapping) to tackle the real-world problem of simultaneously creating an internal map and localizing the robot inside it while the robot explores the environment~\cite{durrant-whyte_simultaneous_2006}. While SLAM can be implemented with low-cost distance sensors (see~\cite{chong_sensor_2015} for a comparison of sensors for SLAM), handcrafted features and traditional grid maps, the approach has more recently been used with visual input \cite{garcia-fidalgo_vision-based_2014}, topological maps \cite{choset_topological_2001} and even Deep Learning algorithms \cite{sotoodeh_bahraini_slam_2019}. We recommend the survey by \cite{fuentes-pacheco_visual_2015} for a detailed introduction to visual SLAM. A combination of Deep Learning, Reinforcement Learning, topological maps and SLAM was recently presented by Chaplot et al. in \cite{chaplot_neural_2020}. 

\paragraph{Socially aware Robot Navigation using Topological maps}
An application of topological maps for socially aware robot navigation was proposed by Johnson et al.~\cite{johnson_socially-aware_2018}, who learn general social behaviour as well as biases (such as moving on the right). A survey on semantic mapping was compiled by Kostavelis et al. \cite{Kostavelis2015}, with an emphasis on human-robot interaction.

%\newline
In contrast to the map-based navigation strategies, mapless or map-free algorithms never explicitly build an internal map of the environment. Early examples such as \cite{matsumoto_visual_1996} employ visual input and template matching, while \cite{santos-victor_divergent_1993} use stereo vision for navigation. Mapless navigation has historically been difficult to evaluate and received less attention than approaches based on the popular SLAM paradigm. Interestingly, with the increased interest in Reinforcement Learning for robot navigation and its combination with Deep Learning, mapless navigation has been a topic of interest again, albeit indirectly - see for example \cite{zhu_target-driven_2017}. The MDP (Markov Decision Process) model, that Reinforcement Learning is based on, does not demand an explicit model or map of the environment and therefore many current navigation algorithms are technically mapless, although they are not explicitly labelled as such. \\
Recent works have also broadened the definition of maps and environment representations. While the environment can be stored in a metric grid, it is hard to reliably create and maintain such a map \cite{engelson_error_1992}. More flexible options are the topological maps mentioned before or the compressed internal representations of Neural Networks \cite{mirowski_learning_2019}. For a comprehensive review of Representation Learning approaches we recommend the work by Bengio et al. \cite{bengio_representation_2013}. A map can also be created purely based on the specific task the robot is trying to solve, as in the work by Zhao et al. \cite{zhao_building_2015} that focuses on a healthcare context. This representation hypothesis has led to the usage of auxiliary tasks related to multi-task learning (see \cite{zhang_survey_2017} for a survey on multi-task learning) that are known to improve another main Deep Learning task in the realm of Reinforcement Learning and Navigation \cite{jaderberg_reinforcement_2016}. It could be argued that this is a return to ``map-based" algorithms with a broader understanding of maps or a middle ground between mapless and map-based.

\subsection{Navigation Tasks}
In this section, we will look at specific tasks that are currently used to evaluate robot navigation algorithms operating in three-dimensional environments. Some of the tasks are derived from specific applications (e.g. a search and rescue robot needs to perform exploration \cite{zelinsky_mobile_1992}), while others are derived from the availability and quality of sensors, such as the tasks in \cite{anderson_evaluation_2018} which require GPS and RGB(D) input. Some of them are more relevant for the academic community than for the deployment of actual physical robots, where human comfort and material wear and tear might play a more important role. The overall goal of all of these tasks is to have a common language for specific problems that can be used for benchmarking in the research stage. To allow for a fair comparison, these benchmarking tasks tend to be evaluated in a simulated environment, which has its own advantages and drawbacks.\\
These tasks are intentionally designed to be simple and straightforward in terms of the problem description, allowing for easy evaluation and comparison. Of course, successfully completing these tasks is just one step in the development of a fully autonomous robot that can assist humans in social, industrial or healthcare contexts. The deployment in social situations and in healthcare environments in particular will require more complex behaviour, but breaking down behaviour into subtasks such as the ones mentioned in this section is still useful if not necessary. For example a doctor could ask an assistive robot to fetch supplies or patients as part of its overall assistive behaviour.
\subsubsection{PointGoal Navigation}

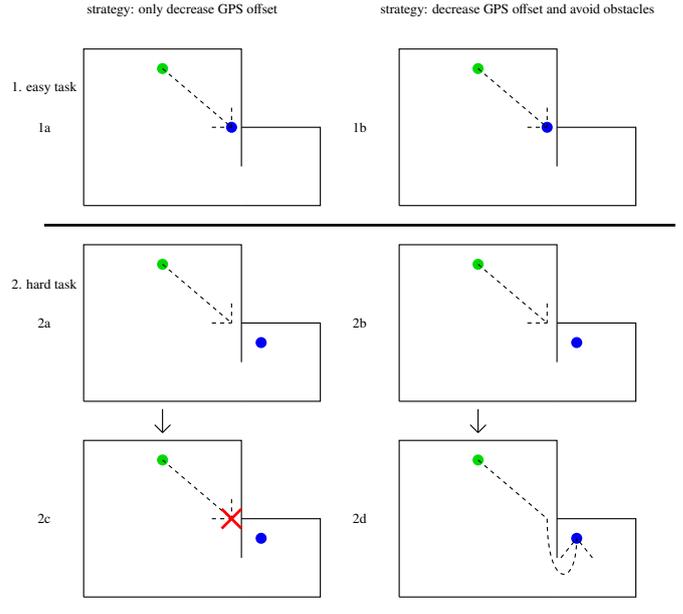
\begin{figure}
\centering
\resizebox{\columnwidth}{!}{%
\begin{tikzpicture}
\draw[black, line width=2] (0,-0.5) -- (16,-0.5);
\draw[dotted]
    (0,2) node {2c};

\draw[black,thick] (1,0) -- (5,0);
\draw[black,thick] (5,1) -- (5,4) -- (1,4) -- (1,0);
\draw[black,thick] (5,0) -- (7,0) -- (7,2) -- (5,2);
\fill[green!85!black] (3,3.5) circle (4pt);
\fill[blue!95!black] (5.5,1.5) circle (4pt);
\draw[black,thick,dashed] (3,3.5) -- (4.75,2);

\draw[black,thick,dashed] (4.25,2) -- (4.75,2);
\draw[black,thick,dashed] (4.75,2.5) -- (4.75,2);

\draw[red, line width=2] (5.0,1.75) -- (4.5,2.25);
\draw[red, line width=2] (4.5,1.75) -- (5.0,2.25);

\draw[dotted]
    (8,2) node {2d};

\draw[black,thick] (9,0) -- (13,0);
\draw[black,thick] (13,1) -- (13,4) -- (9,4) -- (9,0);
\draw[black,thick] (13,0) -- (15,0) -- (15,2) -- (13,2);
\fill[green!85!black] (11,3.5) circle (4pt);
\fill[blue!95!black] (13.5,1.5) circle (4pt);
\draw[black,thick,dashed] (11,3.5) -- (12.75,2);
\draw[black,thick,dashed] (12.75,2) .. controls (12.75,0.2) and (13.5,0.2) .. (13.5,1.5);
\draw[black,thick,dashed] (13.9,1) -- (13.5,1.5);
\draw[black,thick,dashed] (13.1,1) -- (13.5,1.5);

\draw[dotted]
    (0,7) node {2a};
\draw[dotted]
    (0,8) node {2. hard task};

\draw[black, thick] (3,4.2) -- (3,4.8);
\draw[black, thick] (3,4.2) -- (3.2,4.4);
\draw[black, thick] (3,4.2) -- (2.8,4.4);

\draw[black,thick] (1,5) -- (5,5);
\draw[black,thick] (5,6) -- (5,9) -- (1,9) -- (1,5);
\draw[black,thick] (5,5) -- (7,5) -- (7,7) -- (5,7);
\fill[green!85!black] (3,8.5) circle (4pt);
\fill[blue!95!black] (5.5,6.5) circle (4pt);
\draw[black,thick,dashed] (3,8.5) -- (4.75,7);

\draw[black,thick,dashed] (4.25,7) -- (4.75,7);
\draw[black,thick,dashed] (4.75,7.5) -- (4.75,7);

\draw[dotted]
    (8,7) node {2b};

\draw[black, thick] (11,4.2) -- (11,4.8);
\draw[black, thick] (11,4.2) -- (11.2,4.4);
\draw[black, thick] (11,4.2) -- (10.8,4.4);

\draw[black,thick] (9,5) -- (13,5);
\draw[black,thick] (13,6) -- (13,9) -- (9,9) -- (9,5);
\draw[black,thick] (13,5) -- (15,5) -- (15,7) -- (13,7);
\fill[green!85!black] (11,8.5) circle (4pt);
\fill[blue!95!black] (13.5,6.5) circle (4pt);
\draw[black,thick,dashed] (11,8.5) -- (12.75,7);

\draw[black,thick,dashed] (12.25,7) -- (12.75,7);
\draw[black,thick,dashed] (12.75,7.5) -- (12.75,7);

\draw[black, line width=2] (0,9.5) -- (16,9.5);

\draw[dotted]
    (12,15) node {strategy: decrease GPS offset and avoid obstacles};

\draw[dotted]
    (3.5,15) node {strategy: only decrease GPS offset};

\draw[dotted]
    (0,12) node {1a};
\draw[dotted]
    (0,13) node {1. easy task};

\draw[black,thick] (1,10) -- (5,10);
\draw[black,thick] (5,11) -- (5,14) -- (1,14) -- (1,10);
\draw[black,thick] (5,10) -- (7,10) -- (7,12) -- (5,12);
\fill[green!85!black] (3,13.5) circle (4pt);
\fill[blue!95!black]  (4.75,12) circle (4pt);
\draw[black,thick,dashed] (3,13.5) -- (4.75,12);

\draw[black,thick,dashed] (4.25,12) -- (4.75,12);
\draw[black,thick,dashed] (4.75,12.5) -- (4.75,12);

\draw[dotted]
    (8,12) node {1b};

\draw[black,thick] (9,10) -- (13,10);
\draw[black,thick] (13,11) -- (13,14) -- (9,14) -- (9,10);
\draw[black,thick] (13,10) -- (15,10) -- (15,12) -- (13,12);
\fill[green!85!black] (11,13.5) circle (4pt);
\fill[blue!95!black] (12.75,12) circle (4pt);
\draw[black,thick,dashed] (11,13.5) -- (12.75,12);
\draw[black,thick,dashed] (12.25,12) -- (12.75,12);
\draw[black,thick,dashed] (12.75,12.5) -- (12.75,12);
\end{tikzpicture}
}
\caption{ Visualisation of the PointGoal Task in increasing levels of difficulty (from the first row to the last two rows) and with different solution strategies (columns). The agent starts at the green dot and has to move to the blue dot. The black outline denotes the floor plan of a simple environment. }
\label{fig:pointgoal_vis}
\end{figure}

In response to the traction that robot navigation had gained, Anderson et al. \cite{anderson_evaluation_2018} proposed common problem definitions and language to deal with the influx of many different algorithms and evaluation protocols that were for the most part incompatible and incomparable. They propose three main navigation tasks, of which the first, PointGoal navigation, is currently common in publications \cite{mishkin_benchmarking_2019} \cite{kadian_are_2019} \cite{gupta_cognitive_2019}. The other two, ObjectGoal and AreaGoal are slight variations of the first one. The PointGoal navigation task is mastered successfully if the robot navigates to a certain position (point) in the 3D environment from any starting position and orientation (see Figure \ref{fig:pointgoal_vis}).  In the the easy case, where start and stop position are in the same room (row 1), the solution would be trivial - just choose the action that decreases the GPS distance. In a more complex case (the two lower rows), the agent also has to learn how to navigate around obstacles - in this case a wall - to arrive at the destination. While the general goal of decreasing the GPS distance remains valid (2a, 2b), the agent learns to explore in order to deal with obstacles, even if the GPS distance is increased for a short amount of time (2d), otherwise it will fail (2c).\\
The standard PointGoal navigation task assumes that the robot gets GPS information encoding the distance or displacement to the goal as well as other sensors on top if desired. \\

This task definition was also used for the Habitat-challenge\footnote{\url{https://aihabitat.org/challenge/2019/}}, an international challenge that aimed at benchmarking visual robot navigation in 2018. Unfortunately, this specific definition has two obvious drawbacks:
\begin{itemize}
    \item The assumption of a precise displacement sensor is unrealistic in everyday settings, where the robot will have to deal with sensor noise and actuation noise, especially considering the problems indoor environments pose for GPS sensors;
    \item The availability of GPS coordinates for the goal is unrealistic to begin with in most real settings.
\end{itemize}

Independently of the drawbacks, the task does serve as a ``proof of concept" for navigation algorithms if the chosen environments are complex enough in terms of obstacles. Any algorithms that can successfully navigate along the path of the shortest geodesic distance (the shortest path to the goal around obstacles) and does not just blindly decrease the geometric distance (the GPS displacement) has to have learned certain semantic properties of the scene. Although the classic PointGoal navigation task remains a common evaluation baseline it has become obvious that it is hard to translate into the real world. Researchers have started to look into the influence of noise (see the work by Rosano et al. \cite{rosano_embodied_2020} and Chaplot et al. \cite{chaplot_learning_2020}) and task definitions that do not assume the constant availability of the perfect GPS displacement\footnote{\url{https://aihabitat.org/challenge/2020/}}. \\
Very similar iterations of this task are ObjectGoal navigation, where the agent is asked to navigate to an object instead of a point, and AreaGoal navigation, where the agent is asked to navigate to an area.

\subsubsection{Exploration}
In contrast to the previous task, where the exploration of unknown areas of the environment was treated as a byproduct of solving the specific task, some work treats exploration itself as the primary goal. An early approach by \cite{zelinsky_mobile_1992} starts from a task that is similar to PointGoal navigation (with given GPS coordinates for the agent and the goal) and then proposes a customizable exploration strategy that creates an internal topological representation while navigating to the goal.\\
While the mentioned work focuses on tactile sensors, later examples \cite{franz_learning_1998} use images and present a topological map-building algorithm that can be used for exploration. In applications where the exploration is not motivated by a long-term goal, different evaluation methods have been proposed that in turn have shaped different exploration strategies. One common strategy is frontier-based \cite{yamauchi_integrating_1999}, where the robot learns to navigate to the boundaries between explored and unexplored space. Alternatives maximize coverage \cite{stachniss_exploring_2004},\cite{chaplot_learning_2020} to cover areas and recently the authors of~\cite{jayaraman_learning_2018} \cite{ramakrishnan_emergence_2019} minimized uncertainty about the environment in a task-independent approach. In the uncertainty-based approach the agent is asked to predict unseen parts of the environment and the goal is to make this prediction match up with reality in as short a time as possible, which can be applied to robot exploration as well. A recent method by Ramakrishnan et al. \cite{ramakrishnan_occupancy_2020}, the winner of the 2020 Habitat PointNav Challenge, uses an exploration strategy where the robot anticipates a top down occupancy map.\\

\subsubsection{Embodied Question Answering}
This task could be seen as a natural extension of the PointGoal and ObjectGoal navigation task from the previous section. Embodied Question Answering was presented in \cite{das_embodied_2017} as a way to test an algorithm's capabilities under ``natural" circumstances. In this setting, the robot receives a natural language question such as ``What colour is the sofa in the living room ?" and has to navigate in the environment and provide the correct answer in a natural language sentence. The authors of~\cite{gordon_iqa_2018} present a dataset based on AI2-Thor \cite{kolve_ai2-thor_2017} and an algorithm that can answer questions that require interactions such as ``How many apples are in the fridge ?". Wijman et al. present a possible solution with RGB Images and point clouds as input in \cite{wijmans_embodied_2019}.\\
The combination of natural language understanding, exploration, active perception and goal-oriented navigation makes this task very challenging but, unlike PointGoal, it represents an actual use case for embodied agents in domestic settings. Indeed, this task could be integrated into the behaviour of a companion or healthcare robot and make the interaction with a patient or user a lot more natural. \\

\subsection{Popular Paradigms}
\label{Popular Paradigms}
Historically, visual robot navigation has been achieved with classic image processing tools such as edge detectors and optical flow \cite{desouza_vision_2002}, but - like in the field of Image Processing in general - the approaches have changed and evolved into more complex algorithms. A popular choice (for SLAM solutions in particular \cite{durrant-whyte_simultaneous_2006}) is using Gaussian distributions and therefore variations of Kalman filters to track local features and later, non-Gaussian approaches such as particle filters~\cite{montemerlo_fastslam_2003}. \\

Another well-established paradigm is Fuzzy Logic and Control, which is well suited to model imprecise and non-numerical information and decisions. It is a widely used in automation technology \cite{isermann1998fuzzy}, recommendation systems \cite{yera2017fuzzy} as well as medical inference systems \cite{das2016medical} and Human-Robot Interaction \cite{Kuo2014}. \\

The most recent change has come with the revival of Neural Networks and in particular Deep Convolutional Neural Networks \cite{krizhevsky_imagenet_2012}, which now achieve dominating performance, especially in combination with Reinforcement Learning \cite{arulkumaran_brief_2017}. \\

The paradigms mentioned here are not necessarily disjoint and can also be combined - as shown by \cite{Duan2005} who use a fuzzy Reinforcement Learning approach for Robot Navigation, an approach that draws heavily on the fuzzy version of Q-Learning presented by Glorennec et al. \cite{glorennec1997fuzzy}.

\subsubsection{Fuzzy logic and control}
Fuzzy Logic and, as a result, Fuzzy Control, are based on the assumption that certain processes or decisions might be modeled best not with binary states (true or false) but with more imprecise descriptors such as ``warm", ``cold", ``big" or ``small". The idea was formalised by Lotfi Zadeh \cite{ZADEH1965338}, who introduced Fuzzy sets, building on earlier earlier work on many-valued or infinite-valued logic, such as the work by Lukasiewicz \cite{lukasiewicz1920three}. \\
The Process for Fuzzy Control contains 3 main steps: 
\begin{enumerate}
    \item The input data is converted into a state where it is described in terms of memberships to certain ``Fuzzy sets" (for example ``young", ``old", ``middle-aged"), which can be understood as the controller´s version of how a human expert would describe the input.
    \item With the converted - fuzzy - input data , several operations (logical operations such as ``AND", ``OR", ``NOT") are performed to calculate the fuzzy output. The conversion correspont to the ``IF-THEN" rules a human operators would use to justify decisions.
    \item The generated output is then de-fuzzified (often with a weighting scheme) to produce numerical output.
\end{enumerate}
Compared to the more complex solutions, such as Deep Neural Networks, Fuzzy Control schemes have the advantage that they are completely ``readable", since both the definitions of the fuzzy sets and the logical operations on them can easily be transformed into verbal instructions.\\
Fuzzy Control has been employed in the context of robot navigation, see \cite{Saffiotti1997} for an early overview as well as the more recent one by Hong et al. \cite{Hong2012}. Given its reactive nature, Fuzzy Control is useful for collision avoidance: The works by Omrane et al \cite{Omrane2016} and and Pandey et al. \cite{Pandey2014} combine collision avoidance and goal-oriented navigation into a single fuzzy controller, Parhi et al. \cite{parhi2005navigation} not only include obstacle avoidance in cluttered environments but also other robots.

\begin{figure}[t]
\resizebox{\columnwidth}{!}{%
\includegraphics[width=\textwidth]{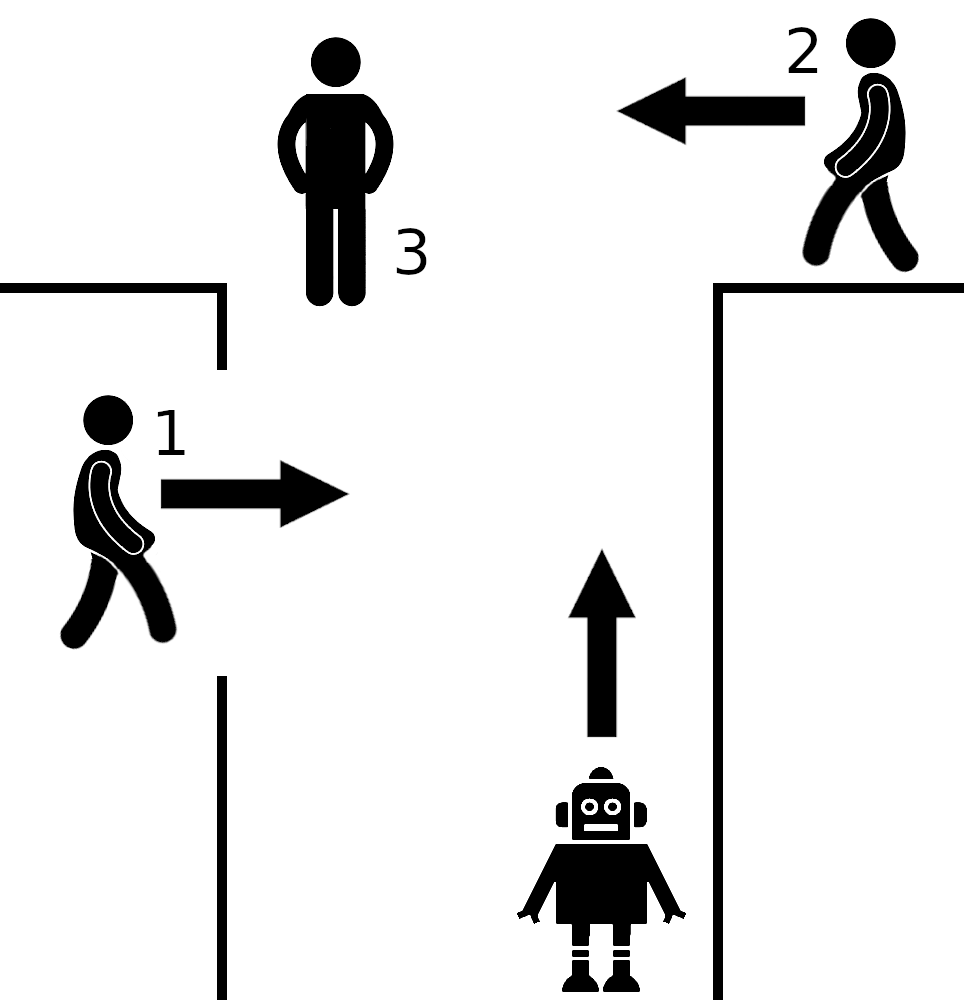}
}
\caption{ Illustration of how a fuzzy controller could be used for socially aware hallway navigation. These prototypical encounters could easily be controlled with if-then rules. 1 - If the human comes form the left, do not yield. 2 - If the human comes from the right, yield. 3 - If the human is stationary, move around. }
\label{fuzzy_figure}
\end{figure}

\paragraph{Socially aware Navigation using Fuzzy Control}
 A solution from the area of Fuzzy Logic was proposed by Palm et al. \cite{Palm2016}, wherein they use Fuzzy Control to predict the movement of humans in relation to the robot, as well as their intentions and desire to compete or cooperate with the robot that shares their space. Fuzzy Control was also used by \cite{Obo2018}, who developed a framework for task planning in crowds. A hallway example can be found in Figure \ref{fuzzy_figure}.\\

\subsubsection{Reinforcement Learning}
Machine Learning can be broadly separated into three major paradigms: 1. Supervised Learning, 2. Unsupervised Learning and 3. Reinforcement Learning \cite{sutton_reinforcement_2018}. The first branch is the ``classic" use case where, based on an existing dataset, the task is to learn the correspondence between input samples and the correct output. Unsupervised Learning, in contrast, explores cases where no labels are available and therefore no input-output correspondence is known. It is more concerned with finding intrinsic structures in the data and inferring similarities and differences from those. Reinforcement Learning, the third branch, uses a trial-and-error approach to learn optimal behaviour of an ``agent" in an ``environment". The aim is less to learn the one correct output or action for a certain input but instead overall optimal behaviour. This makes Reinforcement Learning particularly well-suited for modeling the behaviour of robots (the robot takes the place of the agent and the world it operates in corresponds to the environment) and for training them to perform a certain task. 
We recommend the survey by Arulkumaran et al. \cite{arulkumaran_brief_2017} for an overview on Reinforcement Learning for Robot Navigation. A very recent survey by Akalin et al. \cite{akalin_reinforcement_2020} explores the use of Reinforcement Learning in Social Robotics.

\paragraph{Socially aware Navigation using Reinforcement Learning}
Recent works by Chen et al.~\cite{chen_socially_2017} and Chen et al.~\cite{chen_crowd-robot_2019} suggest that the complex problem of socially aware navigation could be tackled with a holistic end-to-end approach based on reinforcement learning. While the first work focuses on ``what not to do" when navigating in proximity to humans, the second work is more concerned with navigation among crowds. Other solutions that employ reinforcement learning have been proposed by Kim et al. \cite{kim_socially_2016}, Kretzschmar et al. \cite{kretzschmar_socially_2016} and Okal et al.\cite{okal_learning_2016}.

\subsubsection{Learning by Demonstrations}
An intuitive approach to learning behaviour is to learn from a teacher or expert by replicating their actions, also called Learning by Demonstration. A comprehensive overview of Robot Learning from Demonstration was provided by Argall et al. \cite{Argall2009}, who looked at the design strategies, the data acquisition and the various learning and optimisation techniques that can be employed in Learning from Demonstration for robots. A more recent survey was compiled by Ravichandar et al.~\cite{Ravichandar2020}. An example of expert demonstrations that cannot be mapped perfectly to the robot was presented by Skoglund et al.~\cite{Skoglund2010}, who used the arm movements of a human expert to train a robot arm with a different morphology, thus highlighting the importance of generalisation and smart abstraction. In the context of Robot navigation, Learning by Demonstration has been used by Du et al. \cite{Du2019}, who used a combination of group behaviour replication and human following to navigate sidewalks and Xiao et al. \cite{Xiao2020}, who explored the adaption of navigation strategies via human expert demonstrations.\\

\paragraph{Socially aware Navigation using Learning by Demonstration}
Since pedestrians can intuitively respond to  human-like behaviour instead of ``optimal" behaviour, Learning by Demonstration is naturally suited to the training of social robots. Li et al. \cite{li_role_2017} used roleplaying in simulation to train a robot to navigate in a socially compliant manner with Reinforcement Learning.  The demonstrations were mined from real-world trajectory datasets. A shopkeeper robot was trained by Liu et al. \cite{Liu2016DataDrivenHL}, who used examples of human behaviour to train speech and navigation behaviour. Another Learning by Demonstration strategy for social navigation was suggested by Higueras et al. \cite{Higueras2018}, using Inverse Reinforcement Learning and Optimal Rapidly-exploring Random Trees that can be trained with demonstrations. See table \ref{socially_aware_demonstration_table} for a comparison.

\begin{table}[t]
\resizebox{\columnwidth}{!}{%
{\renewcommand\arraystretch{1.25}
\begin{tabular}{|l|l|l|l||l|l|l|l|l|l|l|l|} \hline
\multicolumn{2}{||p{2cm}|}{Publication} & \multicolumn{2}{||p{2cm}|}{problem definition as written in the paper} &\multicolumn{2}{p{2cm}|}{underlying algorithm} & \multicolumn{2}{p{2.5cm}|}{used dataset}  & \multicolumn{2}{p{2.5cm}|}{used metrics}  & \multicolumn{2}{p{2.5cm}|}{tests with real humans}\\ \hline\hline

\multicolumn{2}{||p{2cm}|}{\cite{li_role_2017}} & \multicolumn{2}{||p{2cm}|}{navigate socially with human companion} & \multicolumn{2}{p{2cm}|}{\raggedright Reinforcement Learning}  & \multicolumn{2}{p{2.5cm}|}{\raggedright simulation} & \multicolumn{2}{p{2.5cm}|}{\raggedright terminal condition that encodes: reached goal, collisions, lost companion } & \multicolumn{2}{p{2.5cm}|}{\raggedright yes}\\ \hline

\multicolumn{2}{||p{2cm}|}{\cite{Liu2016DataDrivenHL}} & \multicolumn{2}{||p{2cm}|}{reproducing observed social interactions} & \multicolumn{2}{p{2cm}|}{\raggedright Clustering, Bayesian classifiers}  & \multicolumn{2}{p{2.5cm}|}{\raggedright proprietary dataset} & \multicolumn{2}{p{2.5cm}|}{\raggedright correctness of wording, speech and motion, appropriateness of responses and in social role } & \multicolumn{2}{p{2.5cm}|}{\raggedright yes}\\ \hline

\multicolumn{2}{||p{2cm}|}{\cite{Higueras2018}} & \multicolumn{2}{||p{2cm}|}{learning (social) navigation behaviors} & \multicolumn{2}{p{2cm}|}{\raggedright Inverse Reinforcement Learning }  & \multicolumn{2}{p{2.5cm}|}{\raggedright proprietary dataset} & \multicolumn{2}{p{2.5cm}|}{\raggedright time to goal, path length, distance to people and others} & \multicolumn{2}{p{2.5cm}|}{\raggedright yes}\\ \hline

\end{tabular}}
}
\caption{Comparison of the Algorithms that use Learning by Demonstration for Socially aware Navigation.}
\label{socially_aware_demonstration_table}
\end{table}

\subsection{Simulation Based Learning}
\label{Simulation Based Learning}
Training robots to navigate from scratch purely on real data is notoriously difficult, due to the costs involved in using a real robot, which could be damaged in the early stages of learning. This problem is only exacerbated by the inclusion of humans in the training procedure for interactive navigation. Indeed, it is not feasible or responsible to demand that human participants engage in procedures that might directly harm them in cases of collision. Furthermore, the algorithms that are currently preferred are very ``data-hungry". Deep Learning approaches are known to require very large training sets, the size of which increases the more free parameters the neural networks have - a number that becomes prohibitively big even for simple architectures and downright impossible for more complex architectures if the dataset has to be created manually  (AlexNet, for example, has 60 million parameters \cite{krizhevsky_imagenet_2012}). Reinforcement Learning, with its approach of learning by interacting with the environment again and again further compounds this issue. As a result, it is downright necessary to learn in parallel, which is impossible in the real world but possible in simulations. Natural movement in particular is hard to represent in a classic dataset of samples because the dataset would have to contain the sensor inputs of all possible positions and orientations the robot can be in. In response to this, some datasets restrict the robot movement and only provide samples at certain step sizes or orientations (some datasets circumvent the orientation complexity by providing 360 degree panoramic images for visual input, covering all possible orientations in one sample  \cite{xiao_recognizing_2012}). Examples for this type of dataset are the classic Matterport3D dataset \cite{chang_matterport3d_2017} and the Active Vision dataset \cite{ammirato_dataset_2017}.\\
An alternative to ``classic" datasets, that contain samples, are simulators that are combined with an API which can provide simulated input at every possible position and orientation. These simulators are harder to implement and require more work to provide ``realistic" movement, lighting, etc.. Examples for this approach are Habitat (with API) \cite{savva_habitat_2019}, AI2Thor (with the extensions iThor and RoboThor) \cite{kolve_ai2-thor_2017} and iGibson \cite{xia_interactive_2019}.\\
A common problem with robot navigation is also that, due to the different tasks, different robot setups with their specific types of movement and sensor configurations and the resulting different datasets, it is very complicated and sometimes impossible to compare algorithms fairly. Although many researchers make a point to evaluate their algorithms on a real robot in the real world \cite{deitke_robothor_2020}, this evaluation is usually only qualitative in nature to check if the algorithm transfers properly, but it is not usable as a benchmark. Exceptions to this rule are community challenges, where robots are asked to complete certain tasks or obstacle courses in the same real environment. Examples are the Mohamed Bin Zayed International Robotics Challenge\footnote{\url{https://www.mbzirc.com/}} with a focus on (one or several) robots, the AlphaPilot drone challenge~\cite{foehn_alphapilot_2020}\footnote{\url{https://www.lockheedmartin.com/en-us/news/events/ai-innovation-challenge.html}} and the RoboThor challenge\footnote{\url{https://ai2thor.allenai.org/robothor/challenge/}}, where a robot is trained in a simulated version of a real room and has to navigate to an object in the real room at test time.\\
To combat this issue and in an attempt to guide research, steps have been taken to build simulators that are as universal as possible, with a variety of features that can be used by researchers to reliably benchmark their specific algorithms. As previously discussed, we will be focusing on visual input but for a more general view at simulation environments for robots, we recommend the surveys by Staranowicz et al. \cite{staranowicz_survey_2011} and Torres-Torriti et al. \cite{torres-torriti_survey_2014}.
\subsubsection{Simulation-to-real-transference}
Unfortunately, even with impressive simulation frameworks, some issues remain. No matter how realistic the environment and the robot are modeled they still do not behave like they would in the real world and even small inconsistencies or biases in the simulation setup can skew the results of comparisons or competitions. A great example of this issue has been highlighted by the developers of the Habitat simulator themselves~\cite{kadian_are_2019}. They report that the agents that perform particularly well in the Habitat challenge (using the Habitat simulator), tend to exploit inconsistencies in the simulator, namely a small inaccuracy in the implementation of collisions. The team working on iThor has tackled the problem by creating a physical room with a corresponding counterpart in their simulation \cite{deitke_robothor_2020}. \\
Some solutions to combat issues like the ones above have been suggested. We will highlight some examples in the following:

\textit{Close the gap by making the simulator very realistic:  }

Simulators such as AI Habitat\footnote{\url{https://aihabitat.org/}} \cite{savva_habitat_2019} and Gibson \cite{xia_gibson_2018} use scans of real environments wich are advertised to be photorealistic. iGibson (an interactive extension of Gibson)\footnote{\url{https://github.com/StanfordVL/iGibson}} \cite{xia_interactive_2019} and iThor \cite{kolve_ai2-thor_2017}\footnote{\url{https://ai2thor.allenai.org/}} focus on modelling realistic interaction with dynamic objects in the scene. Habitat has also recently started to push the development of sensor and actuation noise models to minimize the difference between the real world and the simulation that results from noise. Another way to close the gap was proposed by Alhaija et al.  \cite{alhaija_augmented_2017}, who augmented real data with simulated components, thereby guaranteeing variability and a certain level of realism.

\textit{Learn a transfer function from simulated to real data:}

In cases where it is difficult to realistically model the environment the robot should operate in, a possible strategy might be to accept the differences and learn how to cope with them. The problem of translating images from a source domain to a different target domain has been tackled by Zhu et al. with their Cycle-GAN \cite{zhu_unpaired_2017}. Shrivastava et al. \cite{shrivastava_learning_2017} proposed the usage of GANs to train a transfer function that converts a simulated image into one that is indistinguishable from a real photo~\cite{golemo_sim--real_2018}. An example from the domain of robot navigation is the work by Li et al. \cite{li_unsupervised_2020}, who utilize a domain-invariant content space and learn the correspondences between the content space and the source domain as well as the one between the content space and the target domain. The survey by Csurka \cite{csurka_domain_2017} provides a comprehensive overview about visual domain adaptation techniques with a section dedicated exclusively to synthetic-to-real-transference.

\textit{Pretraining on simulated data, finetuning on real data:} 

Since one of the problems of working with synthetic data is that the algorithm will eventually overfit on the synthetic domain, it is intuitive to stop the training process before it is completely finished and then finetune the network on real data. Examples of this strategy can by found for segmentation and recognition problems: Richter et al. \cite{richter_playing_2016} use synthetic data from the video game Grand Theft Auto (a popular game series in the academic community to model high fidelity scenes in the automotive context) to improve performance on the Cambridge-driving Labeled Video Database. Hande et al.\cite{handa_understanding_2016} use simulated indoor environments to improve segmentation results on the NYUv2 and SUN-RGBD sets and \cite{zhu_target-driven_2017} use AI2Thor to pretrain a robot that is later finetuned to operate in a real environment. 

\textit{Randomization:} 

A popular strategy to ensure that a policy transfers well (or at least better) to reality is to vary the properties of the simulation in order to train a policy that is robust to inaccuracies in the simulation. Peng et al. \cite{peng_sim--real_2018} demonstrated this by varying the dynamics of the simulation of a robot arm that is tasked with pushing an object - by varying their simulation properties, they trained a policy that was able to execute the task in a real setup as well. A similar task, namely object localisation and robot grasping was tackled by Tobin et al. \cite{tobin_domain_2017} by varying the textures in the simulated environment. James et al. \cite{james_transferring_2017} use a ``goal corridor" approach in conjunction with a simulation to train a robot to perform tidying actions (picking up an object and placing it somewhere else). They achieve invariance to illumination, clutter and occlusion with domain randomisation in the simulator. It is interesting to note that both Tobin et al. and James et al. do not aim for realism in their randomisation efforts at all. In fact, the visual input of the randomisation strategies is highly unrealistic but it still transfers well to reality.

\subsection{Datasets and Simulation frameworks for Robotic Navigation}
In this section we will present a selection of datasets and simulation frameworks that are commonly used to benchmark visual robot-navigation algorithms. We will not go into detail on datasets that do not allow actual exploration, such as the work~\cite{iuzzolino_virtual--real-world_2018}, where the agent can only move along a predetermined path or datasets and frameworks specifically meant for non-grounded robots and outdoor environments (such as AirSim and Flightmare, developed mainly for drone navigation\footnote{\url{https://github.com/microsoft/AirSim}}\footnote{\url{https://uzh-rpg.github.io/flightmare/}}), but we will list simulators that can be used to train grounded and non-grounded agents, see \cite{kaufmann_deep_2020} who used Gazebo for drone acrobatics.

\subsubsection{Datasets based on the real world}
The advantage of using photos or videos of the real world are obviously their high fidelity and the resulting small simulation-to-real-gap to a potential real world setting. Unfortunately, as discussed, real world data acquisition is costly and concessions in terms of continuous movement have to be made as even a small environment will result in a very big dataset with potentially redundant samples if very fine grained motion is allowed. An alternative are 3D scans (or 3D reconstructions based on photos and videos) that can be traversed freely. Another advantage of these 3D models is the potential to modify the model by adding or removing elements and - depending on the simulator - the option of different lighting/camera/movement configurations. Indeed, the robot should ideally be able to handle all these different conditions and a dataset that only consists of single images might not be able to account for this, or at least only with costly augmentation. Something to keep in mind with 3D scans is that they are not yet perfect when low-cost devices are used to acquire them. While for example the Matterport3D~\cite{chang_matterport3d_2017} and Gibson \cite{xia_gibson_2018} datasets are widely used, many publications mention that only a subset of the available scans is actually chosen for training and testing because reconstruction artefacts can lead to unrealistic behaviour or the simulated robot simply getting stuck. The publication that introduced the Habitat simulator \cite{savva_habitat_2019} has an entire paragraph on this issue in the appendix - the authors manually evaluated every scene in the Gibson dataset and did not use those with significant reconstruction artifacts or overall bad quality. Lastly, both 3D scans and photos do not come with segmentation information so in case this is a task that should be trained for, the dataset will have to be annotated manually. 

\subsubsection{Artificially generated environments}
In contrast to the previously mentioned datasets and frameworks, completely artificial environments are built from scratch by a human designer/programmer \cite{deitke_robothor_2020} \cite{straub_replica_2019}. This makes their creation more laborious but allows to avoid the reconstruction artefacts and inaccuracies of 3D scans. Another positive is that since every object in the scene is known, it is a lot easier to generate semantic maps for segmentation tasks. One very impressive example of this type of dataset is Replica \cite{straub_replica_2019}, which is completely handcrafted but very detailed and near photorealistic, including reflections and material properties. Fortunately researchers are not the only demographic that is interested in high-fidelity simulated environments and computer games have advanced to the point that they can be used to reliably train or pretrain algorithms (see the previously mentioned work by Richter et al. \cite{richter_playing_2016} that utilizes the game Grand Theft Auto as a simulator for human movement in a traffic setting). However, even the most intricate handmade environment will still differ from the reality, just in a different way that environments based on photos or videos of the real world do. 

\subsubsection*{A selection of Datasets and Simulation Frameworks}
\begin{figure}[t]
\resizebox{\columnwidth}{!}{%
    \centering % <-- added
\begin{subfigure}{0.3\textwidth}
  \includegraphics[width=\linewidth]{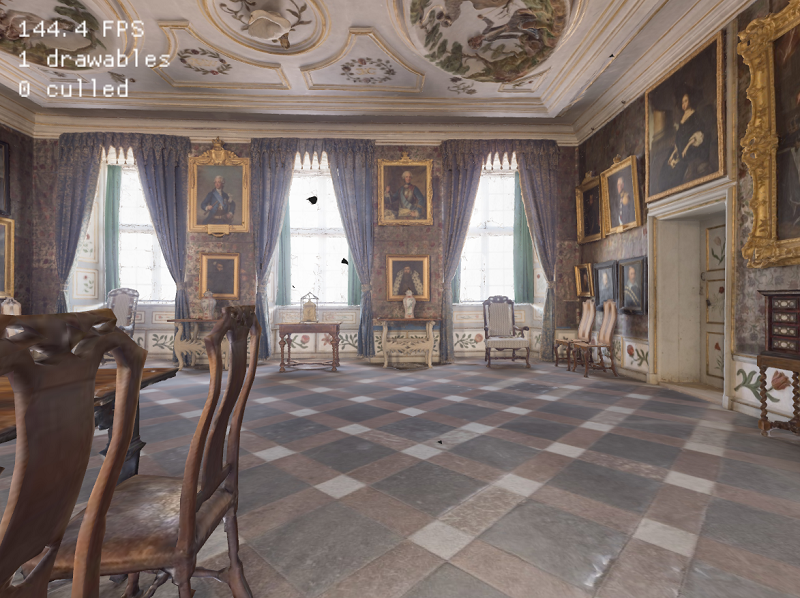}
  \caption{skokloster-castle from Habitat}
  \label{fig:1}
\end{subfigure}\hfil % <-- added
\begin{subfigure}{0.3\textwidth}
  \includegraphics[width=\linewidth]{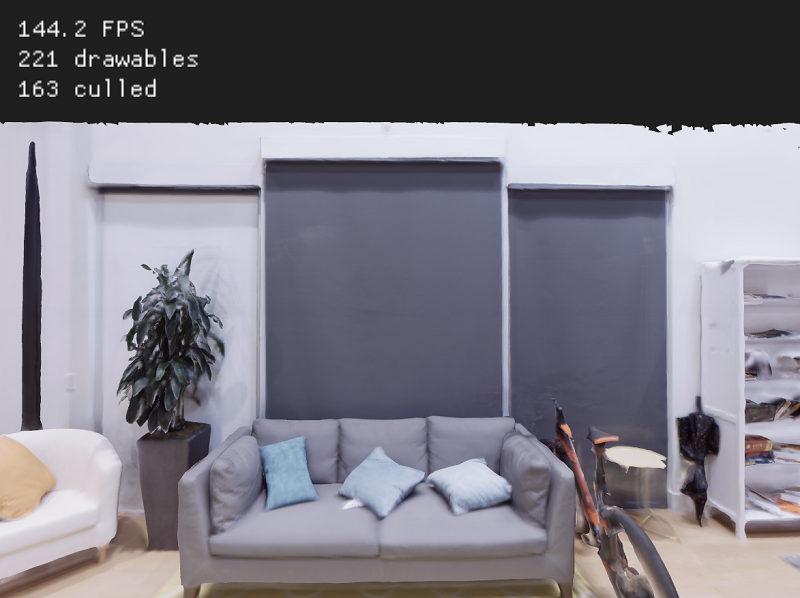}
  \caption{frl\_apartment\_1 from Replica}
  \label{fig:2}
\end{subfigure}\hfil % <-- added
\begin{subfigure}{0.3\textwidth}
  \includegraphics[width=\linewidth]{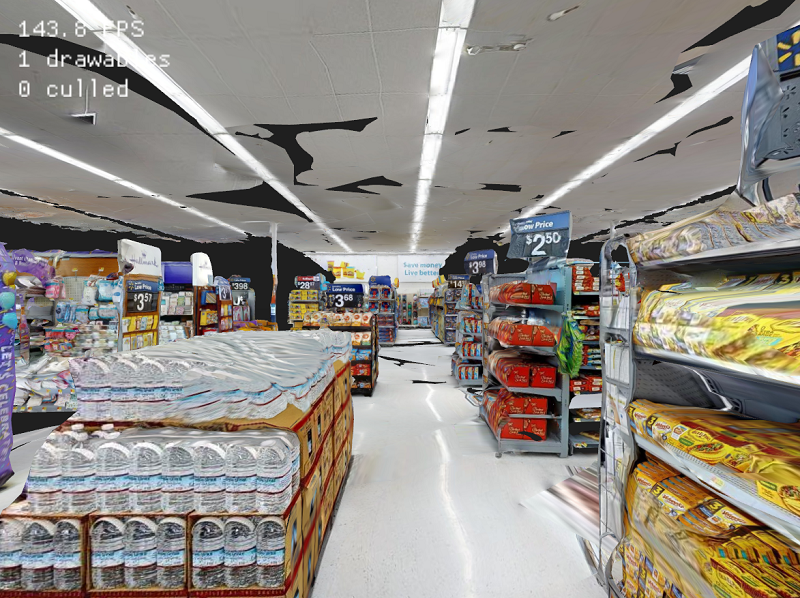}
  \caption{Ackermanville from Gibson}
  \label{fig:3}
\end{subfigure}
}
\resizebox{\columnwidth}{!}{%
\medskip
\begin{subfigure}{0.3\textwidth}
  \includegraphics[width=\linewidth]{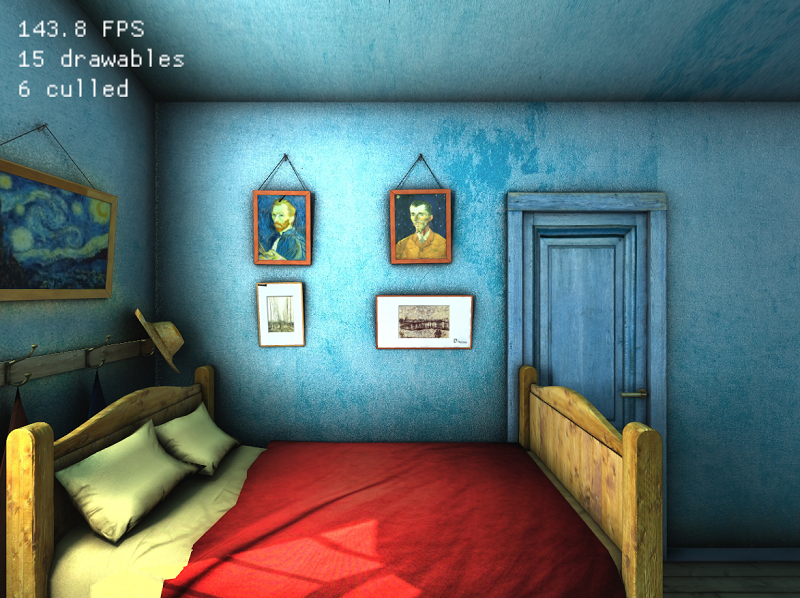}
  \caption{van-gogh-room from Habitat}
  \label{fig:4}
\end{subfigure}\hfil % <-- added
\begin{subfigure}{0.3\textwidth}
  \includegraphics[width=\linewidth]{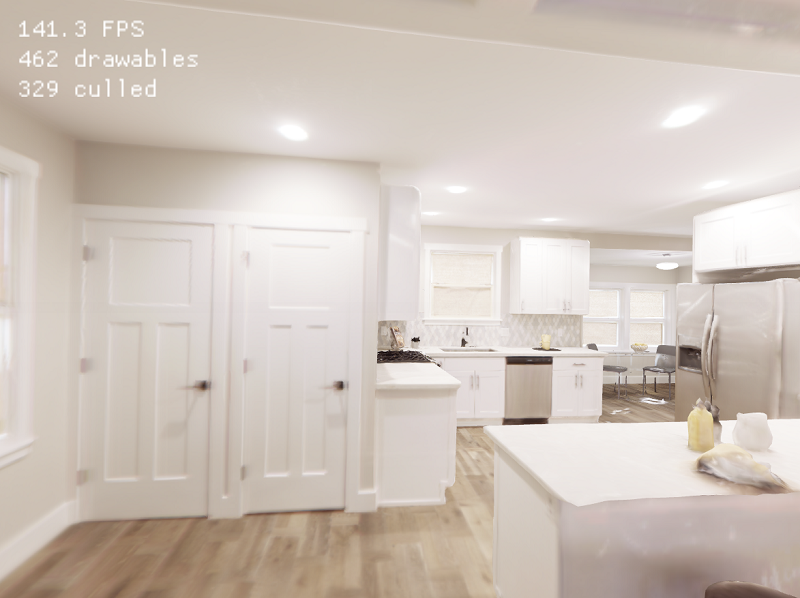}
  \caption{apartment\_0 from Replica}
  \label{fig:5}
\end{subfigure}\hfil % <-- added
\begin{subfigure}{0.3\textwidth}
  \includegraphics[width=\linewidth]{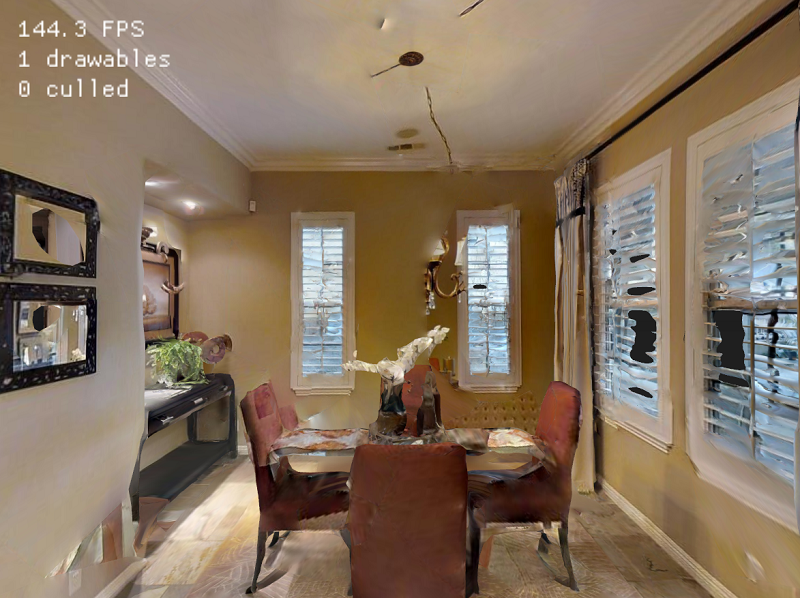}
  \caption{American from Gibson}
  \label{fig:6}
\end{subfigure}
}
\caption{Examples of images generated by Habitat using the demo scenes that come with the simulator, the Replica dataset and the Gibson dataset. The scenes are photorealistic and can be used to create rendered images from every possible viewpoint.}
\label{fig:habitat_imgs}
\end{figure}

We will differentiate between datasets in the ``classic" sense, that consist of discrete data points with or without an underlying structure ,and simulation frameworks that allow continuous movement by rendering the sensor (camera) input for a certain position and orientation on demand. For the purpose of highlighting this difference we will use the term ``framework" whenever an underlying 3D environment is used to generate input data for an agent, although the authors themselves might refer to their work as ``dataset". In contrast, the term ``dataset"  will refer to a collection of images. Some names will come up in both, since it is possible to generate discrete data points from any continuous environment.
\iffalse \textcolor{blue}{In some simulation literature the two types are referred to as query-based (for a dataset comprised of discrete samples) and model-based  REFERENCE.} 
\fi
Figure \ref{fig:habitat_imgs} provides a first impression of the quality of images that can be generated with simulators, showcasing examples from AI Habitat, Gibson and Replica. Tables \ref{sim_table} and \ref{sim_table_2} summarize the main properties of the datasets and frameworks respectively. An interesting ``in-between" approach has been proposed by Rosano et al.~\cite{rosano_embodied_2020}, who used a 3D model of a room in habitat for continuous training, as well as samples from the real room that were registered with the 3D model to provide realistic input.\\
All the datasets and simulators are free, but some require an application process. We will not be listing the SUNCG~\cite{song_semantic_2017} dataset because, even though it has had an impact on the development of robot navigation algorithms, it is no longer available due to a lawsuit of Planner 5D against Princeton and Facebook\footnote{\url{https://futurism.com/tech-suing-facebook-princeton-data}}.\\
\subsubsection*{Classic datasets}

\paragraph{Matterport3D}

The Matterport3D \cite{chang_matterport3d_2017}
%removed references to habitat, removed imprecise descriptors

\footnote{\url{https://niessner.github.io/Matterport/}} dataset is a RGB-D dataset that covers 90 building-sized scenes. The aim of the dataset is to provide seamless view transitions and to this end the dataset includes 10,800 panoramic images generated from nearly 200,000 RGB-D images. It is annotated with segmentation information (2D and 3D) and camera poses, so it can be used for active vision applications as well. Acccording to the authors, the dataset is meant to further research in the topics of keypoint matching, prediction of surface normals, semantic segmentation and region classification. Although the ``original" version of this dataset is comprised of individual samples with an underlying structure, the 3D scans that are used to sample the images can be used for simulation frameworks as well.

\paragraph{Active Vision dataset}
% removed extra info, removed comparison to matterport

The Active Vision dataset\footnote{\url{https://www.cs.unc.edu/~ammirato/active\_vision\_dataset\_website/}}, like the Matterport 3D dataset, is aimed at navigation research in indoor environments and is based on the real world as well. The dataset covers 9 (later extended to 15) unique scenes and contains 20,000+ RGB-D (later extended to 30,000) images with 50,000+ (later 70,000) bounding box object annotations. In their original publication \cite{ammirato_dataset_2017}, the authors demonstrate how the dataset can be used to benchmark algorithms for object instance detection, as well as active vision and navigation. 

\begin{table}
\resizebox{\columnwidth}{!}{%
{\renewcommand\arraystretch{1.25}
\begin{tabular}{|l|l|l|l||l|l|l|l|l|l|l} \hline
\multicolumn{2}{||p{2cm}|}{Dataset} & \multicolumn{2}{p{2cm}|}{Data format} & \multicolumn{2}{p{2cm}|}{Photorealistic} & \multicolumn{2}{p{2.5cm}|}{Interaction with the environment} & \multicolumn{2}{p{3.5cm}|}{Size}\\ \hline\hline

\multicolumn{2}{||p{2cm}|}{Matterport3D \cite{chang_matterport3d_2017}} & \multicolumn{2}{p{2cm}|}{\raggedright RGB-D images} & \multicolumn{2}{p{2cm}|}{\raggedright yes} & \multicolumn{2}{p{2.5cm}|}{\raggedright no} & \multicolumn{2}{p{3.5cm}|}{\raggedright 10.800 panoramic views of 90 multi-floor houses containing 2000 room regions}\\ \hline

\multicolumn{2}{||p{2cm}|}{Active Vision dataset \cite{ammirato_dataset_2017}} & \multicolumn{2}{p{2cm}|}{\raggedright RGB-D images with object bounding boxes} & \multicolumn{2}{p{2.5cm}|}{\raggedright yes} & \multicolumn{2}{p{2.5cm}|}{\raggedright no} & \multicolumn{2}{p{3.5cm}|}{\raggedright 30,000+ RGBD images, 15 scenes}\\ \hline

\end{tabular}}
}
\caption{Table of the classic datasets for robot navigation (samples with underlying structure, no simulation interface).}
\label{sim_table}
\end{table}

\subsubsection*{Simulation frameworks}

\paragraph{UnrealCV}
%removed qualitative descriptors
UnrealCV \cite{weichao_qiu_unrealcv_2017}\footnote{\url{https://github.com/zfw1226/gym-unrealcv}} uses the Unreal game engine and provides researchers with the option to build new environments and interact with them. It can be used in conjunction with deep learning frameworks and to generate ground truth data from a virtual scene. The project website\footnote{\url{https://unrealcv.org/}}  provides documentation and tutorials, including the integration into OpenAI Gym, a framework for Reinforcement Learning. 

\paragraph{MORSE}
% removed popular
MORSE \cite{echeverria_modular_2011}\footnote{\url{https://www.openrobots.org/morse/doc/stable/tutorials.html}} is  a simulator for  academic robotics that focuses on 3D simulation of a variety of environments (small, big, indoor, outdoor). It can simulate one or several autonomous agents and can be accessed with the console or python scripts. It provides a variety of sensors such as cameras, range sensors, GPS and several motion setups. MORSE is based on the Blender engine for visualisation and the Bullet library for physics.

\paragraph{Gazebo}
%removed qualitative descriptors, removed habitat reference

The Gazebo framework\footnote{\url{http://gazebosim.org/}} provides a multitude of features such as test environments, physics simulations, noise models and ready-made virtual representations of real robots (it also supports communication with ROS). On the associated github entry\footnote{\url{https://github.com/osrf/gazebo/tree/master}}, the authors (and the community) offer a selection of the previously mentioned features - this variety in combination with functional features such as cloud support, cross-platform support, command line tools and remote training support makes Gazebo suitable for researchers that are interested in the physical and practical behaviour of robots.\\
As a result, this framework is often used for ``low-level" physics-related research and not so much for ``high-level" conceptual research.

\paragraph{Interactive Gibson}

The Gibson framework, also called iGibson\footnote{\url{http://svl.stanford.edu/igibson/}} to highlight its interactive nature, focuses on interactive navigation. In contrast to previous navigation frameworks, this framework allows the robot to physically interact with the scene and with objects in it. Examples of interactions are: opening drawers or fridges, picking up small items such as fruit and moving slightly bigger objects such as chairs. In terms of size, the framework contains 572 buildings and 1400 floors, as well as several virtual representations of popular robot models (iGibson also supports multi-agent settings). The framework also provides 3D annotations of the object classes chairs, desks, doors, sofas, and tables. The authors of~\cite{xia_gibson_2018},\cite{xia_interactive_2019} introduce the framework in combination with the novel task of navigating in cluttered environments. They present several novel metrics to evaluate the combination of navigation and interaction and apply these metrics to several navigation baselines.\\
This interactive focus sets this framework apart from the others - although some of them have recently started to add interaction options, this framework was especially designed for interaction tasks.

\paragraph{Replica}
%removed qualitative descriptors, removed sim-to-real promise

Replica \cite{straub_replica_2019} is a framework consisting of 18 handmade photo-realistic rooms. The rooms have high-resolution textures with reflection information and class annotations. It was designed for the training of egocentric agents that relies on realistic environments with as few artifacts and simulation biases as possible. Given the information contained in the scenes, Replica can also be used to train algorithms for semantic segmentation in 2D and 3D. The framework comes with a lightweight C++ renderer interface for training\footnote{\url{https://github.com/facebookresearch/Replica-Dataset}}. However, given the affiliation of the authors it is not surprising that even at its introduction, Replica was already designed to be used in Habitat \cite{savva_habitat_2019}.

\paragraph{3RScan and RIO10}

3RScan\footnote{\url{https://waldjohannau.github.io/RIO/}} is a framework for 3D object instance-localization, as in estimating the pose of one or several objects in the same scene that changes over time (for example a living room where books are carried from point A to point B). The authors of \cite{wald_rio_2019} consider this task to be particularly relevant for real world applications - for example when a robot is asked to fetch a specific object that has changed location. The 3RScan framework contains 1482 RGB-D scans of 478 indoor scenes in varying states. For every scan, some objects in the scene have changed position. The scenes are annotated with object class segmentations, as well as the ground truth object position correspondences across the scans. The authors demonstrate a first possible solution to this matching problem with a convolutional 3D correspondence network. Although this framework is primarily meant to explore re-localisation, the high-quality 3D scenes can also be used for active vision and navigation. In the year after 3RScan was introduced, the authors posed an even harder problem with the corresponding RIO10 framework~\cite{wald_beyond_2020}. The purpose of the second framework is to evaluate algorithms that are able to perform re-localisation tasks in changing indoor scenes. In addition to changing objects in the scene, this framework also offers realistic illumination variations. \\
These frameworks explore use cases that are very hard to solve but essential for the deployment of embodied agents in the real world.

\paragraph{Habitat}
% removed a lot of references to habitat and the habitat support, removed qualitative descriptors

Habitat \cite{savva_habitat_2019}\footnote{\url{https://aihabitat.org/}} is a simulation framework developed and maintained by Facebook, with the aim to push the development of embodied navigation. It can be used to evaluate algorithms for active perception, robot navigation and limited interaction scenarios. The framework consists of a simulator and an API that handles the communication between python and the simulator. The repositories\footnote{\url{https://github.com/facebookresearch/habitat-sim}},\footnote{\url{https://github.com/facebookresearch/habitat-lab}} support existing 3D scans from Matterport3D, Replica and Gibson as well as custom environments. In addition, the authors provide baseline implementations of reinforcement learning algorithms such as Actor-Critic \cite{sutton_reinforcement_2018} and PPO \cite{schulman_proximal_2017} as well as DDPPO \cite{wijmans_dd-ppo_2020}. 

\paragraph{AI2 Thor}
%removed qualitative descriptors, removed reference to funding and support
The House Of inteRactions (THOR)\footnote{\url{https://ai2thor.allenai.org/}} \cite{kolve_ai2-thor_2017} is a framework for visual AI research - namely embodied agents. It  contains  30 varied photo-realistic indoor scenes and supports interaction with objects in the scene. AI2 Thor is intended as a starting point for behaviour tasks such as deep reinforcement learning, scene interaction, imitation learning and more classic computer vision tasks such as object detection and segmentation. An  extension  of AI2 Thor is RoboThor \cite{deitke_robothor_2020}, a subset of the original dataset with corresponding real world rooms. The idea is to train robots in simulation and verify their performance in the real world. Another  subset  is iThor, a set of 120 rooms with interactive objects. \\
 The focus on the sim-to-real transfer and the corresponding challenge\footnote{\url{https://ai2thor.allenai.org/robothor/challenge/}} sets THOR apart from other options. 

\paragraph{NVIDIA Isaac NavSim}
% removed qualitative descriptors, removed comparison to habitat and thor
NVIDIA Isaac NavSim\footnote{\url{https://docs.nvidia.com/isaac/isaac/packages/navsim/doc/navsim.html}} is a simulator developed and maintained by NVIDIA that aims at utilizing their GPU functionalities for the simulated training of robots. It is based on Unity3D and provides a variety of test environments as well as tools for automated data generation. NVIDIA Isaac NavSim uses the Isaac SDK that was created specifically to interface between the algorithmic and simulation side (Unity3D, tensorflow) and the used NVIDIA hardware.\\
The desired audience for this simulator is researchers and the industry sector.  Possible use cases are: warehouse interactions (stacking cases, moving objects) and the direct deployment on physical robots. A similar NVIDIA product is the recent NVIDIA Drive Constellation system\footnote{\url{https://developer.nvidia.com/drive/drive-constellation}} for the development of algorithms for self-driving vehicles. 

\paragraph{Unity3Ds mlagents}
%removed qualitative descriptors

\begin{figure}[t]
\resizebox{\columnwidth}{!}{%
\includegraphics[scale=0.5]{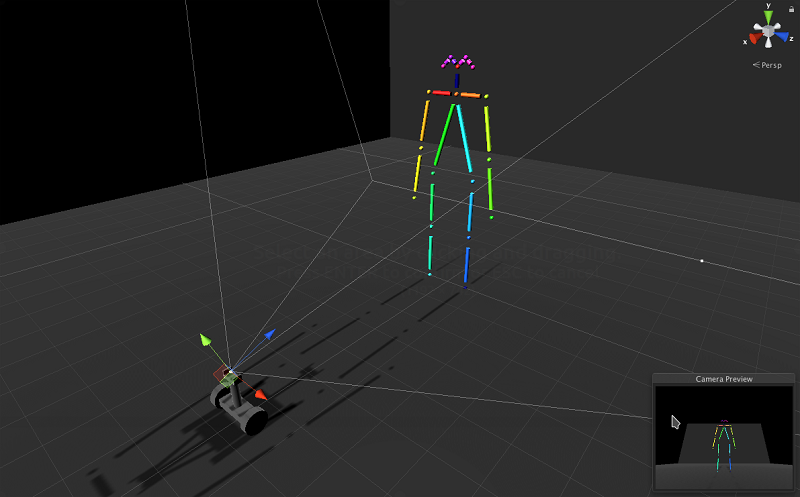}}
\caption{ Example of a very simple simulation setup in mlagents with a small robot and a human skeleton, built in Unity3D. The user can create and customise simulation environments and robot setups. The simulated visual input for the robot is shown on the bottom right. }
\label{simulations_fig}
\end{figure}

Unity3Ds mlagents\footnote{\url{https://unity3d.com/machine-learning}} is a plugin for the Unity3D game engine that is aimed at game developers who want to include artificial intelligence into their games and at researchers who want to build custom environments (Figure \ref{simulations_fig}). It provides the Unity3D functionality for designing and building environments that can be augmented with 3D assets and an interface for setting up Reinforcement Learning agents. It supports different input modalities such as range sensors and images. It is designed primarily for game developers, so the focus is on the generation of the environment, which has made mlagents an entry point into the topic. \\
\begin{table}
\resizebox{\columnwidth}{!}{%
{\renewcommand\arraystretch{1.25}
\begin{tabular}{|l|l|l|l||l|l|l|l|l|l|l|} \hline
\multicolumn{2}{||p{2cm}|}{Dataset} & \multicolumn{2}{p{2cm}|}{data format} & \multicolumn{2}{p{2cm}|}{photorealistic} & \multicolumn{2}{p{2.5cm}|}{interaction with the environment} & \multicolumn{2}{p{3.5cm}|}{size}\\ \hline\hline

\multicolumn{2}{||p{2cm}|}{UnrealCV\footnote{\url{https://unrealcv.org/}}} & \multicolumn{2}{p{2cm}|}{\raggedright simulation framework} & \multicolumn{2}{p{2.5cm}|}{\raggedright yes} & \multicolumn{2}{p{2.5cm}|}{\raggedright no} & \multicolumn{2}{p{3.5cm}|}{\raggedright several available 3d environments, custom environments} \\ \hline

\multicolumn{2}{||p{2cm}|}{MORSE \cite{echeverria_modular_2011}} & \multicolumn{2}{p{2cm}|}{\raggedright simulation framework} & \multicolumn{2}{p{2.5cm}|}{\raggedright no} & \multicolumn{2}{p{2.5cm}|}{\raggedright no} & \multicolumn{2}{p{3.5cm}|}{\raggedright several available environments}\\ \hline

\multicolumn{2}{||p{2cm}|}{Gazebo\footnote{\url{http://gazebosim.org/}}} & \multicolumn{2}{p{2cm}|}{\raggedright simulation framework} & \multicolumn{2}{p{2.5cm}|}{\raggedright no} & \multicolumn{2}{p{2.5cm}|}{\raggedright only between objects in the scene} & \multicolumn{2}{p{3.5cm}|}{\raggedright several available environments, contains creation GUI}\\ \hline

\multicolumn{2}{||p{2cm}|}{Interactive Gibson \cite{xia_interactive_2019}} & \multicolumn{2}{p{2cm}|}{\raggedright simulation framework} & \multicolumn{2}{p{2.5cm}|}{\raggedright yes} & \multicolumn{2}{p{2.5cm}|}{\raggedright yes} & \multicolumn{2}{p{3.5cm}|}{\raggedright  572 buildings, 1400 floors}\\ \hline

\multicolumn{2}{||p{2cm}|}{Replica \cite{straub_replica_2019}} & \multicolumn{2}{p{2cm}|}{\raggedright simulation framework} & \multicolumn{2}{p{2.5cm}|}{\raggedright yes} & \multicolumn{2}{p{2.5cm}|}{\raggedright not in native SDK, yes in AI Habitat} & \multicolumn{2}{p{3.5cm}|}{\raggedright  18 rooms}\\ \hline

\multicolumn{2}{||p{2cm}|}{3RScan \cite{wald_rio_2019}} & \multicolumn{2}{p{2cm}|}{\raggedright simulation framework} & \multicolumn{2}{p{2.5cm}|}{\raggedright yes} & \multicolumn{2}{p{2.5cm}|}{\raggedright no} & \multicolumn{2}{p{3.5cm}|}{\raggedright 1482 RGB-D scans of 478 environments}\\ \hline

\multicolumn{2}{||p{2cm}|}{AI Habitat \cite{savva_habitat_2019}} & \multicolumn{2}{p{2cm}|}{\raggedright simulation framework} & \multicolumn{2}{p{2.5cm}|}{\raggedright yes} & \multicolumn{2}{p{2.5cm}|}{\raggedright yes (some physics support)} & \multicolumn{2}{p{3.5cm}|}{\raggedright supports Replica, Matterport, Gibson and others} \\ \hline

\multicolumn{2}{||p{2cm}|}{AI2 Thor \cite{kolve_ai2-thor_2017}} & \multicolumn{2}{p{2cm}|}{\raggedright simulation framework} & \multicolumn{2}{p{2.5cm}|}{\raggedright yes} & \multicolumn{2}{p{2.5cm}|}{\raggedright yes} & \multicolumn{2}{p{3.5cm}|}{\raggedright 120 rooms (89 apartments for RoboThor)}\\ \hline

\multicolumn{2}{||p{2cm}|}{Nvidia Isaac Sim 2020\footnote{\url{https://docs.nvidia.com/isaac/isaac/packages/navsim/doc/navsim.html}}} & \multicolumn{2}{p{2cm}|}{\raggedright simulation framework} & \multicolumn{2}{p{2.5cm}|}{\raggedright yes} & \multicolumn{2}{p{2.5cm}|}{\raggedright yes} & \multicolumn{2}{p{3.5cm}|}{\raggedright supports custom environments}\\ \hline

\multicolumn{2}{||p{2cm}|}{Unity3Ds mlagents\footnote{\url{https://unity3d.com/machine-learning}}} & \multicolumn{2}{p{2cm}|}{\raggedright simulation framework} & \multicolumn{2}{p{2.5cm}|}{\raggedright yes} & \multicolumn{2}{p{2.5cm}|}{\raggedright yes} & \multicolumn{2}{p{3.5cm}|}{\raggedright supports custom environments}\\ \hline

\end{tabular}}
}
\caption{ Table of the simulation frameworks for robot navigation.}
\label{sim_table_2}
\end{table}

\section{Human Robot Interaction}
Ever since robots have started to act and move, researchers have been interested in studying their interaction with humans, both in social contexts or in work settings where robots and humans have to coexist or even cooperate in the same space. Human Robot Interaction (HRI) as a separate field is relatively young, one of the earliest conferences being the 1992 IEEE International Symposium on Robot and Human Interactive Communication (RoMan). The field is historically multi-disciplinary: researchers from robotics, psychology, natural language processing, cognitive sciences and even philosophy work together and the results are accordingly diverse. Given the increased automation and use of technology in everyday activities, HRI is and will be an important aspect of daily life. \\
In this survey,  we will give a brief introduction of the topic but we will not be focusing on the ethical implications (see~\cite{lin_robot_2011} for a relatively recent survey) or purely psychological work.\\

\subsection{Social Interaction}
The field of social robot interaction studies how humans and robots interact in a natural, ``human-like" manner. Researchers from this field theorize that, depending on the application, humans prefer to interact with robots that learn and display certain social skills \cite{greeff_why_2015}. The long term goal is to create robots that can understand human behaviour, including verbal and nonverbal cues, social conventions and emotional signals and - in order to be intuitively understood by humans - use these tools to communicate information in turn. This dimension of intelligence is considered to be a crucial aspect in designing robots that are a useful and comfortable partner. \\
This way of looking at intelligence runs counter to many paradigms in Machine Learning and Artificial Intelligence, that define intelligence by measurable benchmarks: winning a game of chess (and recently, a game of go \cite{silver_mastering_2017}) against human players, classifying images \cite{krizhevsky_imagenet_2012}, navigating efficiently \cite{savva_habitat_2019} or answering factual questions \cite{das_embodied_2017}. Indeed, social interaction abilities are difficult to evaluate and often subjective. For more comprehensive reading on the topic, we suggest the work by Dautenhahn et al.  \cite{dautenhahn_socially_2007}, \cite{Fong2003} and Breazeal \cite{breazeal_social_2008}, which provide conceptual overviews. Recent works have proposed methods for intention-forecasting in human-robot interaction \cite{ben_amor_interaction_2014} and a reinforcement learning strategy to learn specific social behaviour such as handshakes \cite{qureshi_robot_2016}. On the more holistic side, Triebel et al. \cite{Triebel2016} presented a robot that is able to assist, inform and guide pedestrians in an airport environment. Their system used a combination of detection and tracking, behaviour and social hierarchy analysis and social mapping to generate socially compliant paths for the robot.

\subsection{Interaction in an Industrial Context}
Any work environment that has humans interacting with robots has to evaluate the potential for both significant increase in productivity as well as danger \cite{maurtua_human-robot_2017}. With increased automation, especially in assembly lines, robots are working in closer contact and on more complex tasks with human workers\footnote{for example, see \url{https://secondhands.eu/}}. Early work by Lenz et al. \cite{lenz_joint-action_2008} presented a concept for a collaborative workspace,whereas  Albu-Schäeffer et al. suggested designing a robot specifically for collaborative work \cite{albu-schaeeffer_dlr_2007}. For further reading we point to the recent survey by Tsarouchi et al.\cite{tsarouchi_human_2016}, who review research and progress on HRI in the manufacturing/production environment and the recent work by Villani et al. \cite{villani_survey_2018}.

\subsection{Interaction in the Context of Home Assistance and Healthcare}

\begin{figure}[t]
\resizebox{\columnwidth}{!}{%
\includegraphics[scale=0.5]{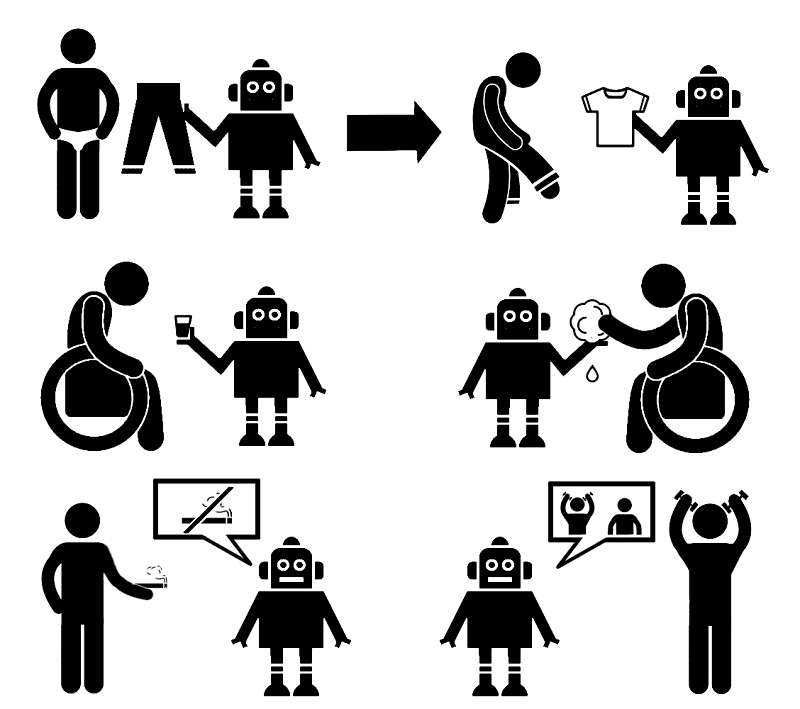}
}
\caption{ Examples of HRI in the context of healthcare - a companion robot can support the patient during daily tasks (putting on clothes, upper row) or take the role of a healthcare professional and help in instances where the patient cannot perform a task alone (second row, getting a drink or bathing with a sponge). Two contactless interactions are shown in the bottom row, namely robot-guided addiction prevention or addiction management and robot-guided exercise. }
\label{healthcare_fig}
\end{figure}

Robots are of big interest in the domain of healthcare and could potentially improve a patient's convalescence in the hospital and at home \cite{dahl_robots_2014} (see Figure \ref{healthcare_fig}). They can be used for observation tasks, assist both doctors and patients and fill care gaps in a time when the average age in developed countries is rising and the healthcare sector is already struggling to find enough qualified staff \cite{fasola_socially_2013}. A robotic companion or ``nurse" can collect information useful for doctors, healthcare workers and the patients themselves \cite{coradeschi_giraffplus_2013}.  The work by Erickson et al. \cite{erickson_multidimensional_2019} tackles both of the tasks of dressing and bathing with an armed robot that can support the patient in real time. Other works proposed robot support in the kitchen \cite{yamazaki_recognition_2011} and robot support for aging citizens in general \cite{yamazaki_home-assistant_2012} . \\

\paragraph{COVID-19}
During the COVID-19 pandemic, robots have been immensely helpful because they can support frontline healthcare workers and reduce the transmission risk. They can also help with auxiliary support tasks such as cleaning and logistics, as well as reconnaissance tasks (for example compliance monitoring during quarantine). The survey by Shen et al. \cite{Shen2021} provides an overview of robotic systems that have been used during the pandemic, from both an industrial and an academic point of view.\\
Given the infectiousness of the virus, it is of vital importance to either use protective gear or avoid contact altogether. Robots can help with both problems. They can monitor patients, provide certain diagnostic features such as taking a temperature, conveying messages and providing instructions. As a result, less healthcare workers were exposed to the virus and less PPE (Personal Protective Equipment) was neccessary\footnote{https://edition.cnn.com/2020/11/11/tech/robots-india-covid-spc-intl/index.html}. Yang et al.~\cite{Yang2020} propose and test a telerobotic system, that combined medical expertise with the adherence to ``social distancing" guidelines. In their review, Seidita et al. \cite{Seidita2020} examined published works and projects in the areas of clinical care, logistics and reconnaissance. \\
On the logistics side, robots can deliver protective material such as PPE, medication or potentially dangerous waste. All of these problems can be tackled with classic goal-oriented navigation strategies. Robots can also disinfect contaminated areas to relieve cleaning personnel and reduce infection\footnote{https://www.lgnewsroom.com/2020/12/lg-announces-autonomous-robot-with-disinfecting-uv-light-for-various-b2b-applications/}\footnote{https://www.phs-uv.com/en/}. Similarly to standard cleaning robots, those robots can employ coverage-based exploration algorithms. \\
On the topic of reconnaissance, robots have been used to monitor patients during quarantine and alleviated isolation in cases where in-person visits would have posed a risk \cite{Seidita2020}. \\

The Covid-19 epidemic of 2020 has also emphasized the role of preventive healthcare \cite{erickson_multidimensional_2019} especially in people who are at a high risk of developing a serious condition and the potential for robots that monitor compliant behaviour, for example social distancing \cite{Sathyamoorthy2020COVIDRobotMS}. \\
\\Beyond Covid-19, there is the on-going epidemic of chronic diseases, which is often considered the largest healthcare problem in the world. Poor lifestyle behaviour is a common factor for the development of various common chronic diseases such as Type II diabetes, and chronic heart disease, and behavior change has been show to be the most effective prevention of these conditions. A companion robot can have an important role in motivating the patient to exercise \cite{Gorer2013} \cite{Lotfi2018}. In home healthcare patients are typically either acutely ill but do not require hospitalization, in rehabilitation after a critical episode, or chronically ill. A robotic assistant may play many roles in these conditions \cite{coradeschi_giraffplus_2013}.\\

In the case of patients with impairments in terms of movement, cognitive abilities, perception, communication or strength, a robot could help by assisting, clarifying, translating, reminding etc.. Unfortunately, the advancements in terms of hardware and software are not yet at the point where robots can be seamlessly integrated into a patient's hospital stay or convalescence at home, but the first promising steps have been taken. The overview by \cite{riek_healthcare_2017} provides a good entry point into the topic of healthcare, identifying tasks, involved parties, challenges and opportunities in this relatively young field. Something to keep in mind in this area is that, even if robots can provably benefit the patient, they still have to be accepted \cite{broadbent_attitudes_2010} and properly used \cite{dawe_desperately_2006} by the patients. The works of Kawamura et al. \cite{kawamura_trends_1994} and~\cite{nemoto_power-assisted_1998} and Fischinger et al. \cite{Fischinger2014} have explored the potential of robots as companions or assistants for the elderly, while the authors of~\cite{tsui_i_2011} and~\cite{tucker_control_2014} propose solutions for disabled patients in the form of mounted wheelchair arms and prosthetics and orthotics, respectively. \\
An interesting subset of research on robots in the context of healthcare is concerned with autistic tendencies. Research has shown that many children with autism spectrum disorders (ASD) respond well to robots and find them easier to communicate with than other humans \cite{scassellati_robots_2012}. A large body of research has covered potential reasons for this effect and how to exploit it to assist children \cite{diehl_clinical_2012}, \cite{begum_are_2016}, \cite{robins_robots_2004}. We especially recommend the works by Scassellati et al. \cite{scassellati_robots_2012} and Cabibihan et al. \cite{cabibihan_why_2013} in this context.

\subsection{Socially aware Navigation for natural Human Robot Interaction}
For a robot to naturally navigate among and interact with humans, it has to not only not endanger humans but also obey a variety of explicit and implicit rules. Some of these rules are very specific, such as a minimal distance to every human to avoid collisions and guarantee a sense of comfort, some are less well-defined such as social conventions (who goes first at an impasse, what side of a path is used) and ``natural" movement. Humans prefer to interact with robots that abide by familiar rules of movement and whose behaviour can be predicted intuitively. Kruse et al. \cite{kruse_human-aware_2013} summarize that the requirements that social navigation strategies have to fulfill are 1. comfort, as in the humans should not be stressed or inconvenienced 2. naturalness, as in the robot should behave similar to humans and 3. sociability, as in the robot should follow cultural conventions. \\
Depending on the underlying task of the robot such as navigating to a certain point, moving with one or several humans or engaging, different solutions have been proposed.\\
A general caveat of all research in this area is that the evaluation of algorithms for Socially aware Navigation is very difficult given its subjective nature. The problem definitions and the resulting datasets vary a lot and given the potential danger, one strategy that is frequently employed is to train on simulated episodes and perform a qualitative evaluation afterwards with real humans. In that case the results will naturally depend a lot on the used simulation framework and the test subjects. \\

\begin{figure}[t]
\resizebox{\columnwidth}{!}{%
\includegraphics[scale=0.5]{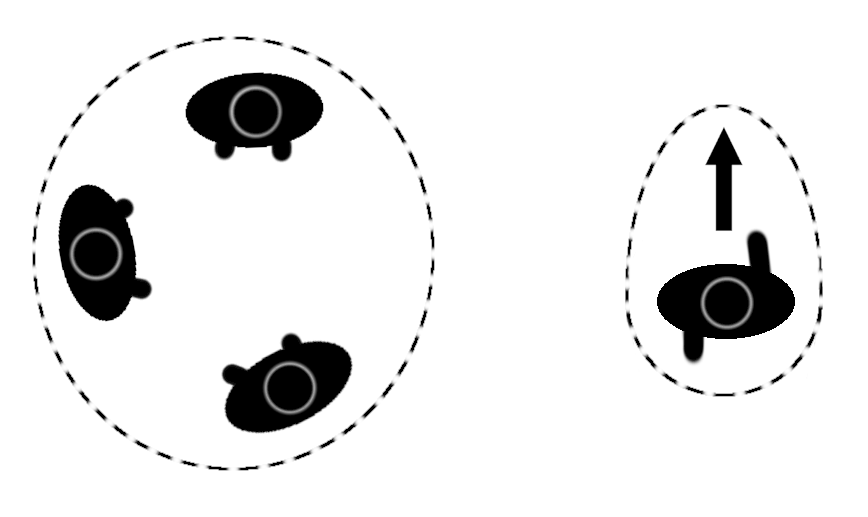}
}
\caption{Illustration of proxemic zones for different activities. On the left, three people are facing each other as a group, the result is a proxemic zone around all of them. On the right, a single person is moving upwards, resulting in bigger zone in the front and a smaller one in the back. The robot should not enter the proxemic zones when it passes.}
\label{proxemic_fig}
\end{figure}

\paragraph{Proxemics}
A big part of guaranteeing behaviour that is perceived as ``safe" is controlling how quickly and how close to the human the robot operates \cite{pacchierotti_evaluation_2006}. The area of research that deals with the amount of space humans like to keep between themselves and others is called Proxemics and is of particular relevance for robots that operate in social situations or in healthcare contexts. Of course this area depends on the individual as well as the social and cultural context, but it is an aspect of human-robot-cohabitation that has to be kept in mind to ensure that the robot can execute its task smoothly, without inconveniencing the human, see Figure \ref{proxemic_fig}. We will present a select few works that explore proxemics for robots, the overall field is much broader and leans heavily into social and cultural studies.\\
The work by Henkel et al. \cite{henkel_evaluation_2014} evaluates different distance strategies by how they affect the human's perception of the robot's likeability, intelligence and submissiveness, Kim et al.~\cite{kim_how_2014} looked at the influence of an underlying task (cooperation vs competition) on the perception. Takayama et al. explored the causes for the difference in perception (such as previous experience with robots or pets)  \cite{takayama_influences_2009}, Obaid et al. \cite{obaid_stop_2016} conducted posture experiments and Mead et al. \cite{mead_perceptual_2016} focused on the design implications of social conventions for human-robot-interactions. While the previous works tended to focus on one-on-one interaction with the robot, Mavrogiannis et al.~\cite{mavrogiannis_effects_2019} explored the effects on a crowd of people.\\
Pacchierotti et al. \cite{Pacchierotti2005} presented a proxemics-based control strategy for hallway navigation and Koay et al. \cite{Koay2014SocialRA} explored proxemic preferences for handover tasks. Papadakis at al. \cite{Papadakis2013SocialMO} model proxemic zones depending on human-human interactions to inform their path-planning strategy.

\section{Analyzing and Modeling Human Behaviour}
For a robot to naturally interact with humans it is crucial to equip them with the ability to not only detect them but also to infer what the human is doing. Different activities might indicate that movement is imminent, that the human does not want to be disturbed or - depending on the task of the robot - that an intervention is necessary. \\
Activity recognition has been evaluated in the context of Human Robot Interaction \cite{Ji2019} as well as the contexts of socially aware path planning and navigation and healthcare - see Farie et al. \cite{faria2015probabilistic} and their follow-up \cite{Vieira2016}.\\
In the rest of this section we will highlight more general Solutions and mention instances where activity recognition is used to inform robot behaviour in the later subsection on Socially aware Navigation using Activity Recognition.\\

The problem of Human Activity Recognition (HAR) is widely studied \cite{jobanputra_human_2019}, due to its varied applications such as surveillance \cite{taha_human_2015}, healthcare \cite{liu_human_2016} and - most relevant for this survey - human-robot interaction~\cite{chrungoo_activity_2014} . Video-based activity recognition is very appealing since data acquisition is relatively cheap and unlabeled data is easy to come by online. Labelling the data, on the other hand, tends to be expensive. Given its complex nature, HAR tends to combine the newest advances in classic computer vision algorithms such as background modeling, motion estimation (optical flow) and stereo vision with more complex machine learning advances such as human pose detection, object-detector neural networks, relation graphs and recurrent neural networks (since actions have a temporal dimension). For a detailed overview on the topic, we recommend Moeslund et al. \cite{moeslund_survey_2006} for the early 2000s and Zhang et al. \cite{zhang_review_2017} for more recent works.\\
Human movement has a very hierarchical nature that has to be taken into account when evaluating it for certain goals. At the lowest level, humans perform motion primitives \cite{zhang_motion_2012} such as moving a hand, turning the torso etc. and these primitives are executed sequentially or in parallel to perform a certain action (in some publications a difference is made between actions and activities, we will use them interchangeably in this survey). A set of actions - again in sequence or in parallel - generates behaviour. A concrete example would be: moving the hand to grab a tomato, coring the complete tomato, cooking a bowl of soup. As per the example, it is not enough to just evaluate the parts of the human that are in motion or still, one also has to take into account what objects are interacted with, either directly (opening a drawer) or indirectly (running after a ball in a game of soccer). Further complexities are introduced with multi-person activities \cite{zinnen_multi_2009} and multi-modal activity recognition (e.g. audio-visual activity recognition \cite{kazakos_epic-fusion_2019}). We will give a brief introduction into core concepts and topics that are of interest to current activity recognition research.
Human behaviour understanding is mostly performed on videos shot from a third person perspective, where most or all parts of the human body are in frame, such as the Kinetics \cite{kay_kinetics_2017} dataset, the ActivityNet \cite{fabian_caba_heilbron_activitynet_2015} dataset and the Charades \cite{sigurdsson_hollywood_2016} dataset. Other datasets have also focused on first person vision, such as EPIC-KITCHENS\footnote{\url{https://epic-kitchens.github.io/2020-100}} \cite{damen_rescaling_2020} and the Georgia Tech Egocentric Activities\footnote{\url{http://ai.stanford.edu/~alireza/GTEA\_Gaze\_Website/}} dataset \cite{fathi_learning_2012}. A robot can use onboard sensors and wearable sensors attached to the human as well as other third-position sensors.\\

\subsection{Pose Estimation}

\begin{figure}[t]
\resizebox{\columnwidth}{!}{%
\includegraphics[scale=0.35]{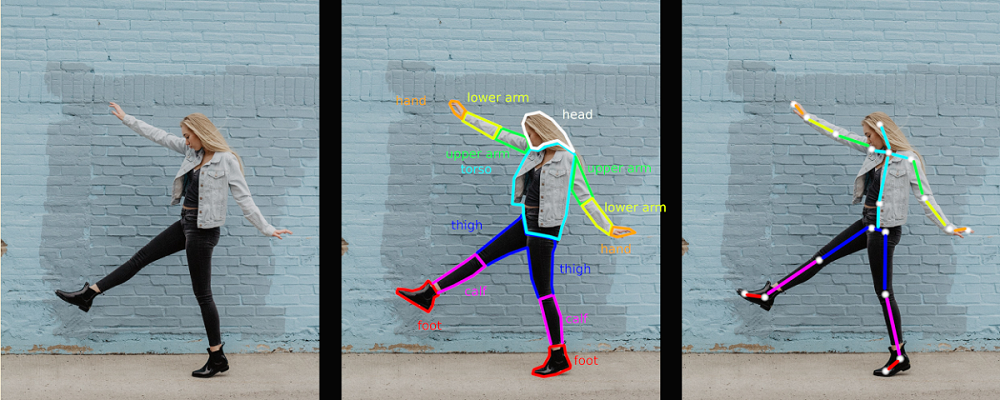}
}
\caption{Main goals of a 2D Human Pose Estimation approach. Starting from the RGB image as input (left - photo by Joanna Nix Walkup on unsplash), algorithms can segment relevant body parts (center) and infer the 2D skeleton of the human considering a given set of joints (right).}
\label{pose_estimation_img}
\end{figure}

The first step in many Human Behaviour understanding pipelines is to estimate where the human is in the frame and how their limbs are positioned, see Figure \ref{pose_estimation_img}. This has the dual benefits of not only extracting the most relevant information for further evaluation, but also significantly compress the information in the frame, which makes further processing easier. Context information such as backgrounds/scenes and or objects can be extracted separately. Although Pose Estimation is well-explored to the point that robust cross-platform solutions such as Open-Pose\footnote{\url{https://github.com/CMU-Perceptual-Computing-Lab/openpose}} \cite{cao_openpose_2019} \cite{simon_hand_2017} \cite {cao_realtime_2017} \cite{wei_convolutional_2016} exist, it remains a well investigated task. Indeed, recent works focus on further refining the exact 2D or 3D position of human joints in image (\cite{tompson_joint_2014} \cite{yasin_dual-source_2016} \cite{bulat_human_2016} \cite{newell_stacked_2016} \cite{wei_convolutional_2016}  \cite{carreira_human_2016} \cite{sun_compositional_2018} \cite{luvizon_human_2019} \cite{chu_multi-context_2017}) and video data (\cite{pfister_flowing_2015} \cite{pavllo_3d_2019}).  Depending on the desired goal, it might also be beneficial to focus only on selected limbs such as the hand, see \cite{Palm2006}, who used  Fuzzy Clustering and Takagi Sugeno (TS) modeling to recognize and imitate grasping behaviours. 
More recent examples  use texture \cite{pavlakos_texturepose_2019}, common sense rules and external information such as objects \cite{yao_modeling_2010} and scene information to finetune further. The recent detectron2\footnote{\url{https://github.com/facebookresearch/detectron2}} library, developed and maintained by facebook, includes several pose estimation modules as well - it was designed and is used as a starting point for more complex evaluations such as action recognition. \\
Since robots (and other surveillance applications) will have varying viewpoints and might not always see the observed human in the ``optimal" way. Indeed, it is of great interest for researchers to develop algorithms that can handle incomplete and occluded input.

\subsection{Action Recognition}
Action recognition from RGB video or images is a very popular research topic that has been tackled in a variety of ways. Just like in other areas of computer vision, early solutions employed low-level vision features such as motion, contours and keypoints \cite{bobick_recognition_2001}  while more recent work tends to favor Deep Learning architectures \cite{kong_human_2018}.  \\
Due to the complex spatiotemporal nature of human action (and its frequent overlap), researchers rarely attempt to solve the overall action recognition problem and instead break down the problem into specific subtasks such as temporal localisation \cite{shou_temporal_2016}, multiview action recognition \cite{wang_cross-view_2014} or co-occurrence~\cite{singla_recognizing_2010}, \cite{helaoui_recognizing_2011}. Alternatively, some researchers focus on speed~\cite{yu_fast_2015}, robustness \cite{wang_robust_2015} or particular feature encodings~\cite{peng_action_2014} \cite{peng_multi-region_2016}. As mentioned in the previous section, pose estimation is a standard preprocessing step and several researchers have presented algorithms that can infer the type of action based on 2D or 3D joint positions \cite{lo_presti_3d_2015}, \cite{shi_two-stream_2019}, \cite{li_actional-structural_2019}, \cite{si_attention_2019}, \cite{vemulapalli_human_2014}, \cite{yan_spatial_2018}, \cite{liu_spatio-temporal_2016}. In addition to skeleton data, human action is also determined by the type of objects that are interacted with \cite{koppula_learning_2012}, \cite{yao_recognizing_2012}, \cite{prest_explicit_2013}. For more detailed overviews we recommend the surveys by Zhang et al. for ``classic" solutions before 2011 \cite{aggarwal_human_2011} and the more current follow ups by Herath et al. \cite{herath_going_2017} and Zhang et al. \cite{zhang_comprehensive_2019}.\\
\subsection{Action Forecasting}

\begin{figure}[t]
\resizebox{\columnwidth}{!}{%
\includegraphics[scale=0.5]{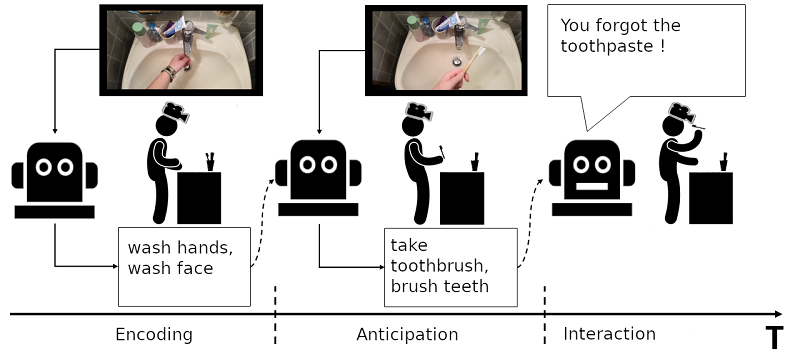}
}
\caption{ Visualisation of the action forecasting problem for the companion robot use case. The robot receives the first person video from a device the human is wearing and first encodes the current action, anticipates possible next actions and plans actions accordingly.}
\label{action_anticipation_fig}
\end{figure}

Another important ability for robots to understand human behaviour is the prediction of future actions, be it from past trajectories or videos, still frames or audio. Being able to anticipate the next action of a human has a wide range of use cases, it is relevant for applications such as automotive safety \cite{jain_recurrent_2015} as well as observation tasks for healthcare, surveillance or social robots \cite{ryoo_robot-centric_2015},\cite{koppula_anticipating_2015}. It is also a very conceptually relevant topic, since any algorithm that can successfully predict the next motion necessarily needs an understanding of how human action is structured and how it might intervene with secondary actions, other agents or objects. Another useful task in this context is trajectory forecasting, which we will touch on later - in some cases the distinction is hard to make, since walking trajectories can be considered actions, like in the early work by Kitani et al. \cite{kitani_activity_2012}. We will list some successful solutions to the different iterations of this problem.\\
Huang et al. \cite{huang_action-reaction_2014} look at human interaction, whereas Gao et al. \cite{gao_red_2017} train a Reinforced Encoder-Decoder (RED) network to predict possible future sequences. In contrast to the majority of works that focuses on the immediate future actions happening in a few seconds, Farha et al. \cite{farha_when_2018} present an algorithm suited for long term prediction. \\
A specific subtask of action forecasting is first-person action prediction, which relies on egocentric input.  Figure \ref{action_anticipation_fig} gives an idea of the possible applications. While the majority of current algorithms and evaluations are geared towards third person action recognition and anticipation, the increasing popularity of head-mounted smart devices such as Google Glass\footnote{\url{https://www.google.com/glass/start/}} and HoloLens2\footnote{\url{https://www.microsoft.com/en-us/hololens}}, combined with the demand for personalised entertainment, support in industry and healthcare applications, could place more emphasis on first-person research in the future. Recent work by Rhinehart et al. \cite{rhinehart_first-person_2017} uses inverse reinforcement learning, while the approach by Furnari et al. \cite{furnari_what_2019} utilizes LSTMs to predict future actions in a domestic setting.

\subsection{Socially aware Navigation using Activity Recognition}
\begin{figure}[t]
\resizebox{\columnwidth}{!}{%
\includegraphics[scale=0.5]{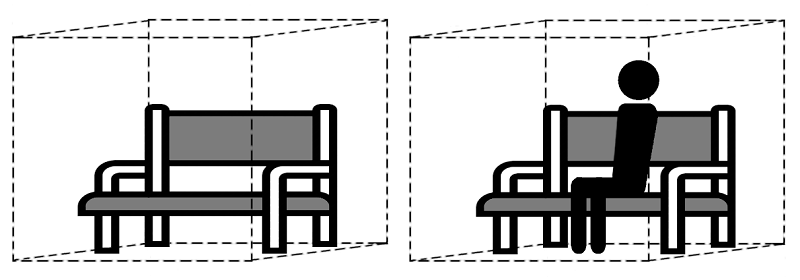}
}
\caption{ Illustration of Affordance (left) and Activity (right) Spaces, denoted by dotted lines, for a bench. Both spaces should ideally not be traversed and definitely not blocked by a passing robot. }
\label{activity_affordance_fig}
\end{figure}

The inclusion of Activity information can be implemented in very basic ways such as slightly improved proxemics-based collision avoidance but it can also include fine-grained activity classes that demand specific behaviour from the robot (for example in a healthcare context \cite{faria2015probabilistic}). The survey by Martinez et al. \cite{Martinez2014} gives an overview of human-aware robot navigation that touches on Proxemics, social conventions and formal definitions for Activity and Affordance Spaces. They suggest that, in addition to avoiding collisions with humans performing certain activities, socially aware robots also have to take into account both Activity Spaces (where an activity is performed) as well as Affordance Spaces (which denote potential Activity spaces, for example a sign than can be looked at), see Figure \ref{activity_affordance_fig}. \\

An example for a comparatively low-level approach was given by Banisetty et al. \cite{Banisetty2016}, who focused on ways the interpersonal distance can be used to infer certain activities and in turn how this distance can be used by the robot to replicate natural behaviour. Robot path planning can also use the information about the performed activity to simply determine what areas should not be traversed by the robot. These areas can be estimated by combining the position of the human and the action that is performed \cite{Magro2018} (see also \cite{Martinez2014} for the definition of Activity and Affordance Spaces) as well as the shape of the interpersonal zone that corresponds to certain actions \cite{Clavero2021}. The action running, for example, corresponds to an elongated forward-facing interpersonal zone that should not be blocked by the robot, see also Figure \ref{proxemic_fig}.\\
Charalampous et al. \cite{CHARALAMPOUS2016261} on the other hand use a more complex approach where human activity is determined by a deep learning module. The information about the performed activity is combined with rule-based proxemics to continuously plan and re-plan the best collision-free path for the robot. The authors also took care to evaluate the algorithm in a human-free setting. An approach focused on group activities was proposed by Okal et al. \cite{Okal2014}. See Table \ref{socially_aware_activity_table} for an overview.\\
\paragraph{Activity recognition for robot navigation}
Given the specific requirements of robotic sensing, Activity recognition has been tackled specifically in this context, with Gori et al. \cite{Gori2015}, who used RGB(D) data to encode visual, velocity and distance features and an SVM and Olatunji \cite{Olatunji2018}, who employed motion capture data and a Convolutional Neural Network to determine the performed activity. An application from the area of domestic robots that includes fine grained Activity Recognition can be found in the work of Piyathilaka et al. \cite{Piyathilaka2013}, who use 3D skeleton features to estimate what action a human is performing to accordingly adjust the behaviour of a robot. The example they give is that a cleaning robot would be able to use information about the current human activity to plan what area to clean next. Rezazadegan et al. \cite{rezazadegan_action_2017} suggested that standard Activity Recognition approaches have to be modified to account for background cues and camera motion before they work on a robot. To this end, they generate action region proposals and then extract shape and motion features to classify the activity with a Convolutional Neural Network. Faria et al. \cite{faria2015probabilistic} used  Spatio-temporal 3D skeleton features to infer not only the performed activity but also potentially dangerous situations such as falls for robot-assisted living.\\
\begin{table}[t]
\resizebox{\columnwidth}{!}{%
{\renewcommand\arraystretch{1.25}
\begin{tabular}{|l|l|l|l||l|l|l|l|l|l|l|l|} \hline
\multicolumn{2}{||p{2cm}|}{Publication} & \multicolumn{2}{||p{2cm}|}{problem definition as written in the paper} &\multicolumn{2}{p{2cm}|}{underlying algorithm} & \multicolumn{2}{p{2.5cm}|}{used dataset}  & \multicolumn{2}{p{2.5cm}|}{used metrics}  & \multicolumn{2}{p{2.5cm}|}{tests with real humans}\\ \hline\hline

\multicolumn{2}{||p{2cm}|}{\cite{Banisetty2016}} & \multicolumn{2}{||p{2cm}|}{appropriate behavior for a mobile robot} & \multicolumn{2}{p{2cm}|}{\raggedright Gaussian Mixture Model}  & \multicolumn{2}{p{2.5cm}|}{\raggedright proprietary dataset} & \multicolumn{2}{p{2.5cm}|}{\raggedright accuracy of activity prediction} & \multicolumn{2}{p{2.5cm}|}{\raggedright yes}\\ \hline

\multicolumn{2}{||p{2cm}|}{\cite{Magro2018}} & \multicolumn{2}{||p{2cm}|}{human-aware navigation} & \multicolumn{2}{p{2cm}|}{\raggedright  clustering, probabilistic roadmap, random tree path planners, elastic band algorithm}  & \multicolumn{2}{p{2.5cm}|}{\raggedright simulation (RoboComp)} & \multicolumn{2}{p{2.5cm}|}{\raggedright time and path length to target, human activity interruption} & \multicolumn{2}{p{2.5cm}|}{\raggedright yes}\\ \hline

\multicolumn{2}{||p{2cm}|}{\cite{Clavero2021}} & \multicolumn{2}{||p{2cm}|}{social navigation for escorting and following} & \multicolumn{2}{p{2cm}|}{\raggedright adaptive proxemics shapes (based on Gaussian functions)}  & \multicolumn{2}{p{2.5cm}|}{\raggedright simulation} & \multicolumn{2}{p{2.5cm}|}{\raggedright mean escorting distance} & \multicolumn{2}{p{2.5cm}|}{\raggedright no}\\ \hline

\multicolumn{2}{||p{2cm}|}{\cite{CHARALAMPOUS2016261}} & \multicolumn{2}{||p{2cm}|}{navigation in a human populated environment} & \multicolumn{2}{p{2cm}|}{\raggedright Deep Learning, rule-based decision making and topological maps}  & \multicolumn{2}{p{2.5cm}|}{\raggedright proprietary dataset} & \multicolumn{2}{p{2.5cm}|}{\raggedright fraction of safe trajectories} & \multicolumn{2}{p{2.5cm}|}{\raggedright yes}\\ \hline

\multicolumn{2}{||p{2cm}|}{ \cite{Vieira2016}} & \multicolumn{2}{||p{2cm}|}{robot path planning with human daily activity recognition} & \multicolumn{2}{p{2cm}|}{\raggedright SLAM and Dynamic Bayesian Mixture
Model}  & \multicolumn{2}{p{2.5cm}|}{\raggedright proprietary datase} & \multicolumn{2}{p{2.5cm}|}{\raggedright Precision and Recall (Activity Recognition)} & \multicolumn{2}{p{2.5cm}|}{\raggedright yes}\\ \hline

\multicolumn{2}{||p{2cm}|}{ \cite{Okal2014}} & \multicolumn{2}{||p{2cm}|}{activity recognition for navigation  in  human  crowds.} & \multicolumn{2}{p{2cm}|}{\raggedright Histograms with velocity features}  & \multicolumn{2}{p{2.5cm}|}{\raggedright simulator} & \multicolumn{2}{p{2.5cm}|}{\raggedright Accuracy (Activity Recognition)} & \multicolumn{2}{p{2.5cm}|}{\raggedright no}\\ \hline

\end{tabular}}

}
\caption{Comparison of the Algorithms that use Activity Recognition for Socially aware Navigation. It is apparent that there is no consistent task definition, dataset or evaluation metric. }
\label{socially_aware_activity_table}
\end{table}

\subsection{Datasets}
In this section we will cover some of the more popular datasets for pose estimation and action recognition. The datasets each have different scopes and unique properties. We will first give a brief overview of each dataset and later summarize the main characteristics in a table for quick comparison. We will group the datasets roughly by the main purpose (Human Pose Estimation and Action Recognition) and whether or not they are based on recordings of real humans.

\subsubsection{Datasets for Human Pose Estimation}
Table \ref{pose_table} summarizes the main datasets for Human Pose Estimation.

\begin{figure}[t]
\resizebox{\columnwidth}{!}{%
\includegraphics[scale=0.5]{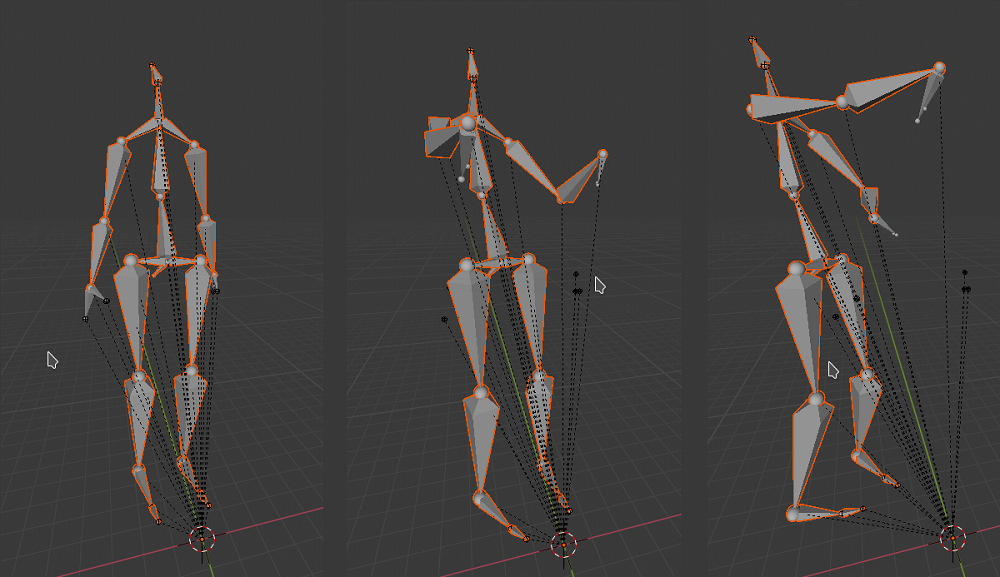}
}
\caption{ Example of 3D Poses from the Human3.6M dataset, visualised in Blender.}
\label{armatures}
\end{figure}

\paragraph{Human3.6M Dataset}
The Human3.6M Dataset\footnote{\url{http://vision.imar.ro/human3.6m/description.php}} \cite{ionescu_human36m_2013} is currently the biggest dataset of human 3D poses collected with motion capture technology and video. It contains 3.6 million 3D human poses and the corresponding videos, as well as 3D meshes of the actors that performed the actions (11 professional actors - 6 male, 5 female). The dataset covers daily and mostly stationary indoor activities such as talking on the phone, smoking, discussing, taking photos etc.. The authors introduce the dataset in \cite{ionescu_human36m_2013} for the training and evaluation human pose estimation models and algorithms. In addition they also provide code for baseline pose prediction algorithms, visualisation and evaluation tools on their website. Three example poses can be seen in Figure \ref{armatures}.

\paragraph{HumanEva}
This dataset is similar to the Human 3.6M dataset, but smaller in size. It is a combination of two datasets, namely HumanEva I and HumanEva II, and like Human3.6M it focuses on indoor environments \cite{sigal_humaneva_2010}. The dataset\footnote{\url{http://humaneva.is.tue.mpg.de/}} consists of RGB/gray-scale videos with synchronized joint positions acquired with motion capture technology. The 7 video sequences contain 4 subjects performing 6 actions such as walking and gesturing.

\paragraph{Total Capture Dataset}
The Total Capture dataset\footnote{\url{https://cvssp.org/projects/totalcapture/TotalCapture/}} \cite{trumble_total_2017} contains around 1.9M frames of indoor multi-view video, IMU and motion capture sensor data. The recordings were made in a very controlled setting using a black motion capture suit and in an empty indoor volume, with 8 cameras, capturing synchronized video streams. The aim of the dataset is to benchmark 3D pose estimating using IMU and multiview cameras. The ground truth poses are provided by the motion capture. In contrast to previous datasets that focus mostly on everyday activities, this dataset also includes more challenging and one-person activities performed by the 5 subjects such as yoga, crawling and bending over.

\paragraph{SURREAL (Synthetic hUmans foR REAL tasks) Dataset}
This dataset\footnote{\url{https://www.di.ens.fr/willow/research/surreal/data/}}  \cite{varol_learning_2017} is meant to provide synthetic data, which can be used to augment real recordings and contains rendered images created from 3D sequences of motion capture data and varying backgrounds. Given how difficult it is to collect and label high quality large scale data for human pose estimation, the dataset can be used to combine random poses, lighting conditions, camera positions and backgrounds - making the most out of existing data (the poses are sampled from the previously mentioned Human3.6M dataset). The authors took several measures to ensure that the generated images are still as realistic as possible. They used the SPML model from \cite{bogo_keep_2016} to create realistic human shapes. Their results demonstrate how the generated dataset improves 2D pose estimation and depth estimation on real images.

\paragraph{JTA Dataset}
The JTA (Joint Track Auto)\footnote{\url{https://github.com/fabbrimatteo/JTA-Dataset}} \cite{fabbri_learning_2018} dataset was created for pose estimation and tracking of pedestrians in urban scenarios. It is rendered using the video game Grand Theft Auto V, a photo-realistic action-adventure game that takes place in an urban environment. The dataset contains 512 full-HD videos (split into training and testing 50/50), each 30 seconds long, recorded at 30 fps. Since the scene is rendered from a known 3D model, synchronized reliable 3d and 2d joint information is automatically available for every video. In addition to the dataset, the authors provide a tool to generate additional videos.

\paragraph{GTA-IM Dataset}
Like the JTA dataset, the GTA Indoor Motion dataset (GTA-IM)\footnote{\url{https://github.com/ZheC/GTA-IM-Dataset}} \cite{cao_long-term_2020} has been created with the video game Grad Theft Auto V. In contrast to JTA, however, its main concern is with indoor human-scene interaction such walking through rooms with furniture, ascending stairs and walking through doorways. The HD RGB-D videos are synchronized with the 3D joint positions and camera pose annoations. The ``actors" in the scenes, as well as the scenes themselves and the actions performed in them are diverse. The creators of this dataset use it to predict how a human will move given an observed trajectory and they find that their results outperform previous approaches both in therms of quantity and quality.
\\
\begin{table}
\resizebox{\columnwidth}{!}{%
{\renewcommand\arraystretch{1.25}
\begin{tabular}{|l|l|l|l||l|l|l|l|l|l|l|l|} \hline
\multicolumn{2}{||p{2cm}|}{Dataset} & \multicolumn{2}{||p{2cm}|}{main research purpose} &\multicolumn{2}{p{2cm}|}{data format} & \multicolumn{2}{p{2.5cm}|}{annotation format}  & \multicolumn{2}{p{2.5cm}|}{size}  & \multicolumn{2}{p{2.5cm}|}{synthetic or real recordings}\\ \hline\hline

\multicolumn{2}{||p{2cm}|}{Human3.6M Dataset} & \multicolumn{2}{||p{2cm}|}{pose estimation}  & \multicolumn{2}{p{2cm}|}{\raggedright videos, joint positions, joint angles, meshes} &  \multicolumn{2}{p{2.5cm}|}{\raggedright activity class, ground truth human pose} & \multicolumn{2}{p{2.5cm}|}{\raggedright 3.6 million 3D poses} & \multicolumn{2}{p{2.5cm}|}{\raggedright recordings of real humans} \\ \hline

\multicolumn{2}{||p{2cm}|}{HumanEva (2009)} & \multicolumn{2}{||p{2cm}|}{pose estimation}  & \multicolumn{2}{p{2cm}|}{\raggedright videos, joint positions} & \multicolumn{2}{p{2.5cm}|}{\raggedright activity class, ground truth human pose} & \multicolumn{2}{p{2.5cm}|}{\raggedright 40,000 frames} & \multicolumn{2}{p{2.5cm}|}{\raggedright recordings of real humans}\\  \hline

\multicolumn{2}{||p{2cm}|}{Total Capture} & \multicolumn{2}{||p{2cm}|}{pose estimation} &\multicolumn{2}{p{2cm}|}{\raggedright videos and joint positions} & \multicolumn{2}{p{2.5cm}|}{\raggedright activity class, ground truth human pose} & \multicolumn{2}{p{2.5cm}|}{\raggedright 1.9 million frames} & \multicolumn{2}{p{2.5cm}|}{\raggedright recordings of real humans}
\\ \hline

\multicolumn{2}{||p{2cm}|}{SURREAL}  & \multicolumn{2}{||p{2cm}|}{pose estimation} & \multicolumn{2}{p{2cm}|}{\raggedright rgbd videoframes} & \multicolumn{2}{p{2.5cm}|}{\raggedright ground truth human pose, segmentations} & \multicolumn{2}{p{2.5cm}|}{\raggedright 1.9 million frames} & \multicolumn{2}{p{2.5cm}|}{\raggedright synthetic and recordings of real humans}
\\ \hline

\multicolumn{2}{||p{2cm}|}{JTA Dataset}  & \multicolumn{2}{||p{2cm}|}{pose estimation} & \multicolumn{2}{p{2cm}|}{\raggedright rgbd videoframes}  & \multicolumn{2}{p{2.5cm}|}{\raggedright ground truth human pose} & \multicolumn{2}{p{2.5cm}|}{\raggedright 256 minutes} & \multicolumn{2}{p{2.5cm}|}{\raggedright synthetic}
\\ \hline

\multicolumn{2}{||p{2cm}|}{GTA-IM Dataset}  & \multicolumn{2}{||p{2cm}|}{pose estimation and trajectory forecasting} & \multicolumn{2}{p{2cm}|}{\raggedright rgbd videoframes} & \multicolumn{2}{p{2.5cm}|}{\raggedright ground truth human pose} & \multicolumn{2}{p{2.5cm}|}{\raggedright 1 million frames, 119 clips} & \multicolumn{2}{p{2.5cm}|}{\raggedright synthetic}
\\ \hline

\multicolumn{2}{||p{2cm}|}{CMU Panoptics dataset} & \multicolumn{2}{||p{2cm}|}{activity recognition and pose estimation} & \multicolumn{2}{p{2cm}|}{\raggedright videos, 3D pose and face landmarks} &  \multicolumn{2}{p{2.5cm}|}{\raggedright activity annotations, speaker iD, ground truth human pose} & \multicolumn{2}{p{2.5cm}|}{\raggedright 65 sequences (5.5 hours)} & \multicolumn{2}{p{2.5cm}|}{\raggedright recordings of real humans}
\\ \hline

\end{tabular}}
}
\caption{Datasets for human pose estimation}
\label{pose_table}
\end{table}

\subsubsection{Datasets for Activity Recognition}
We review a selection of the main datasets here. For a more complete list of datasets, we suggest the overview website by the University of Bonn\footnote{\url{http://actionrecognition.net/index.html}}, which also contains result metrics from state-of-the-art-methods on the listed datasets. The table \ref{activity_table} summarizes the main characteristics of the discussed datasets.

\paragraph{Kinetics }
Kinetics \cite{kay_kinetics_2017} is an extension of the DeepMind Kinetics human action dataset and contains 700 classes of activities. For each activity, the dataset provides 600 or more video clips sourced from youtube. The clips include manually annotated human-object activities and as human-human activities. The website~\footnote{\url{https://deepmind.com/research/open-source/kinetics}} also lists previous iterations of the datasets as well as the relevant publications which contain baseline results. An interesting fact about this dataset is that due to its origin (youtube) it does not exist as one big downloadable dataset, instead the authors provide annotations and instructions on how to download the respective videos. As a result, the actual size of the dataset varies with the chosen video quality and with the availability of each annotated video on youtube. The advantage of this dataset is its impressive size, whereas a drawback is the non-trivial downloading method.

\paragraph{Carnegie Mellon University Multimodal Activity (CMU-MMAC) Database }
The CMU-MMAC database\footnote{\url{https://www.ri.cmu.edu/publications/guide-to-the-} \url{carnegie-mellon-university-multimodal-activity-cmu-} \url{mmac-database/}} \cite{torre_guide_2008} is a multimodal dataset focused on kitchen-related activities. It contains footage of activities such as making brownies, pizza, sandwich, salad and scrambled eggs. The authors used 6 RGB-D cameras with varying spatial and temporal resolutions, one of these cameras is worn by the actor and hence provides a first person view. Additional modalities include sound and motion capture. Although this dataset is relatively small, the varied input modalities differentiate it from other datasets.
\\

\paragraph{ActivityNet}
ActivityNet \cite{fabian_caba_heilbron_activitynet_2015} contains videos of varying human activities, ranging from simple to complex. It covers 203 activity classes with around 140 videos per class and often more than one action per video. The annotations contain global information about the videos, as well as start and stop information for certain activities. The videos are sourced from the internet and annotated using Amazon Mechanical Turk. All in all, the dataset consist of 849 hours of annotated footage. The webiste\footnote{\url{http://activity-net.org/}} provides an introduction, instructions for usage and first results on the dataset for video classification, trimmed activity classification as well as activity detection. The annotations are provided in a hierarchical structure (for example playing sports - discus throw). Like the Kinetics dataset, ActivityNet is sourced from youtube videos that have to be downloaded by the user - the authors provide instructions and code to this end, as well as an evaluation server. Just like with the Kinetics dataset, downloading the dataset is non trivial, but it provides a representative sample size.

\paragraph{Charades and Charades-Ego}
Charades\footnote{\url{https://prior.allenai.org/projects/charades}} and Charades-Ego\footnote{\url{https://allenai.org/plato/charades/}}~\cite{sigurdsson_hollywood_2016} are datasets of daily activities, recorded from a first and, in the case of Charades-Ego, also a third person perspective. For the creation of these datasets, the actors (through Amazon Mechanical Turk) were asked to film themselves performing a certain activity (similar to the game charades, hence the name). The results are varied datasets of 9848 videos with 66,500 temporal annotations for 157 action classes (Charades) and 7860 videos with 68,536 temporal annotations for 157 action classes (Charades-Ego). Since the videos were recorded by the actors themselves, these datasets are very  diverse, but the setup of the data acquisition process (acting out something) makes them not as natural as others.
\\
\paragraph{Cornell Activity Datasets}
The CAD-60 and CAD-120 datasets\footnote{\url{https://www.re3data.org/repository/r3d100012216}} \cite{sung_human_2011} consist of 60 and 120 RGB-D video sequences of four humans performing activities. The available modalities are RGB video, depths information and tracked skeletons. They were recorded with a Kinect sensor and cover activities (such as brushing teeth, cooking, relaxing), each containing sub-activities. The four different actors were recorded in different daily indoor environments, such as a kitchen, living room, office, etc. The inclusion of depth and skeleton data sets these datasets apart from the previous ones and makes them very interesting also for pose estimation, although they are much smaller.
\\

\begin{figure}[t]
\resizebox{\columnwidth}{!}{%
    \centering % <-- added
\begin{subfigure}{0.3\textwidth}
  \includegraphics[width=\linewidth]{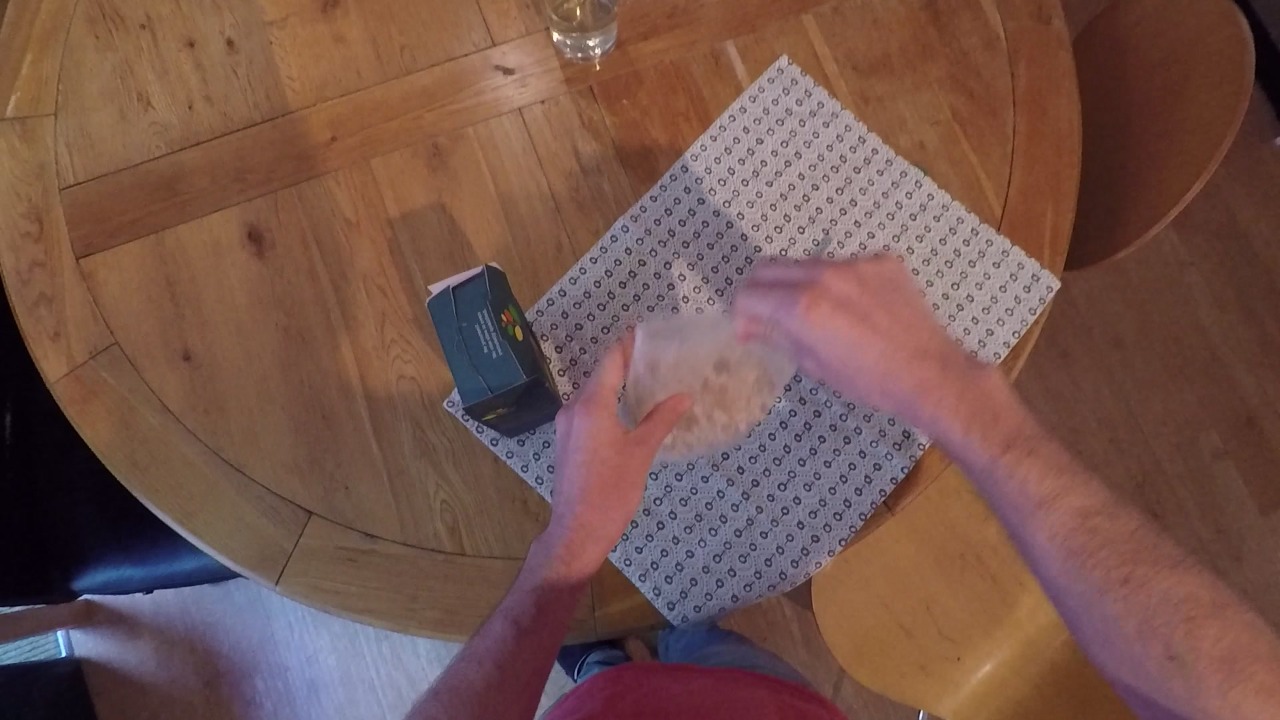}
\end{subfigure}\hfil % <-- added
\begin{subfigure}{0.3\textwidth}
  \includegraphics[width=\linewidth]{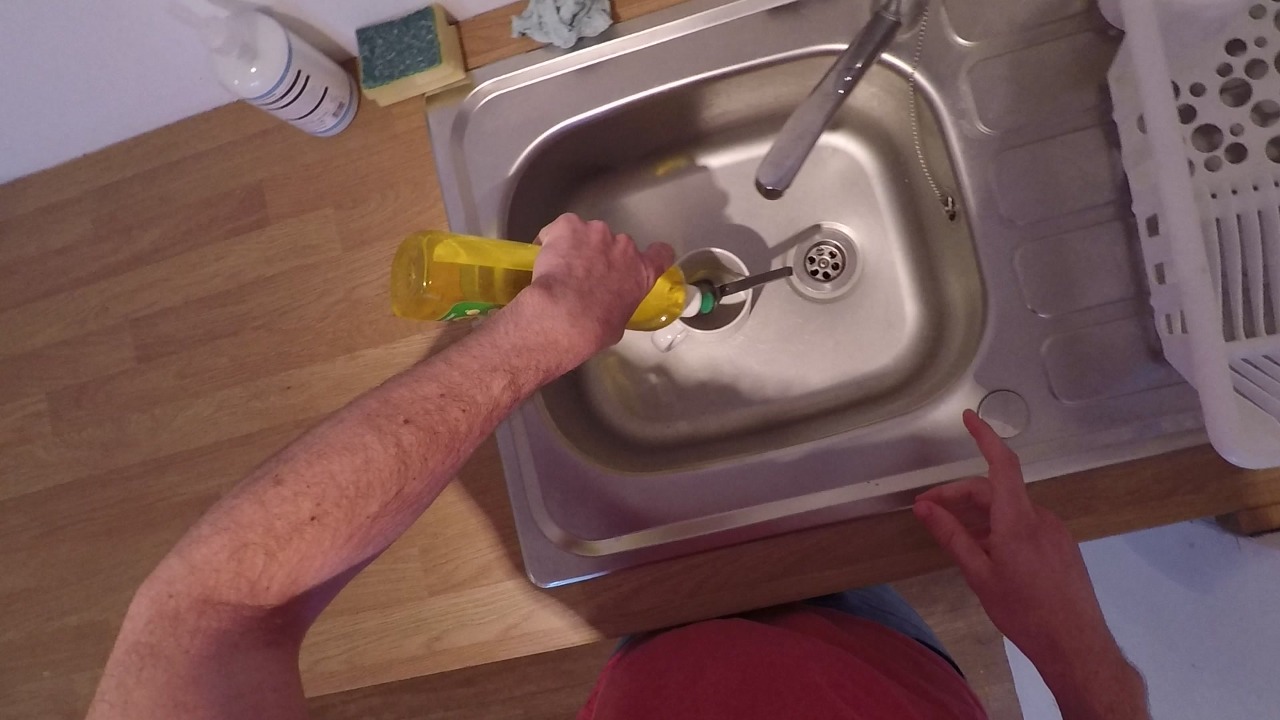}
\end{subfigure}\hfil % <-- added
}
\resizebox{\columnwidth}{!}{%
\medskip
\begin{subfigure}{0.3\textwidth}
  \includegraphics[width=\linewidth]{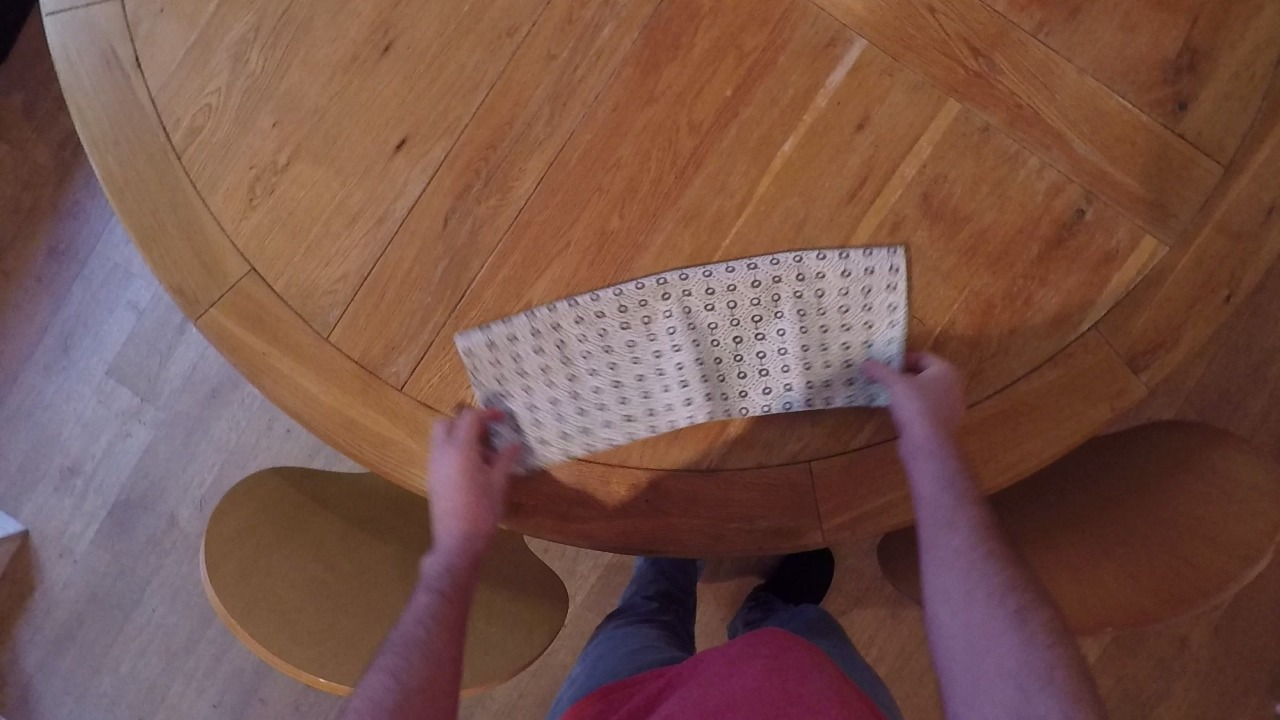}
\end{subfigure}\hfil % <-- added
\begin{subfigure}{0.3\textwidth}
  \includegraphics[width=\linewidth]{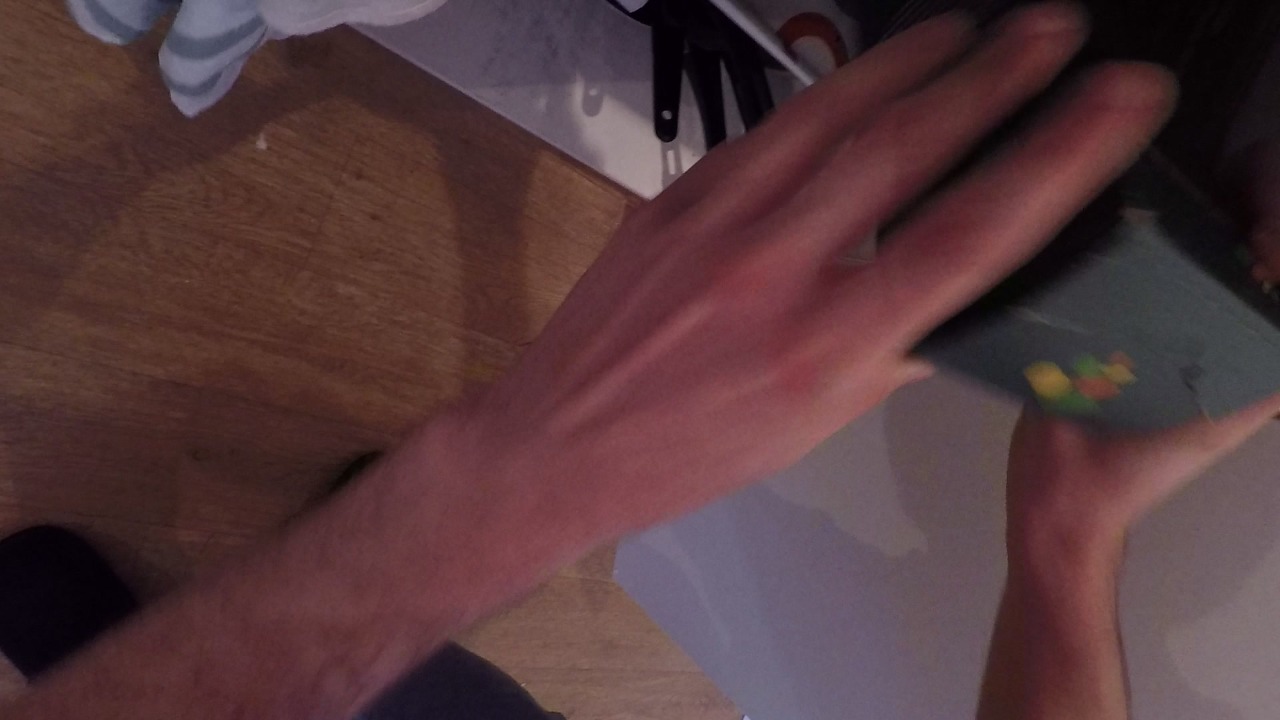}
\end{subfigure}\hfil % <-- added
}
\caption{ Example frames from the EPIC KITCHENS dataset \cite{damen_rescaling_2020}.}
\label{fig:epic_kitchens}
\end{figure}

\paragraph{EPIC-KITCHENS}
EPIC-Kitchens-100\footnote{\url{https://epic-kitchens.github.io/2020-100}} \cite{damen_rescaling_2020} is the extension of the existing EPIC-KITCHENS-55 (for 55 hours) \cite{damen_scaling_2018}, a dataset focusing on egocentric videos of daily activities that occur in the kitchen, such as cooking, making drinks, cleaning etc. (see Figure \ref{fig:epic_kitchens}). The actions are non-scripted, performed in the actors' homes and densely annotated via narration, temporal action segments and object bounding boxes. The extended dataset contains 100 hours of footage with 90.000 action segments, 20.000 unique narrations, 90 verb classes and 300 noun classes. It was published in conjunction with set of challenges to push the research in the area of egocentric vision. The variety, size and scope of this dataset make it very attractive for researchers who are specifically focused on egocentric vision. While other datasets contain some egocentric videos, this dataset is geared completely towards it.

\paragraph{Extended Georgia Tech Egocentric Activity Gaze+ Dataset}
Georgia Tech offers several datasets for egocentric activity recognition\footnote{\url{http://cbs.ic.gatech.edu/fpv/}}, the most recent one is EGTEA Gaze+ \cite{li_eye_2018}. This dataset consists of HD videos with the corresponding audio, gaze trajectories, hand mask segmentations and frame-level annotations for the activities that are performed. The 32 subjects in the dataset performed cooking activities in 86 sessions, amounting to 28 hours of footage. In total, the dataset contains 10325 action instances and 15,176 hand masks on 13,847 video frames.

\paragraph{ADL (Activities of Daily Living) Dataset}
This dataset \cite{pirsiavash_detecting_2012} of activities was recorded with a chest-mounted camera that recorded HD videos of 18 daily living activities performed by 20 actors\footnote{\url{https://www.csee.umbc.edu/~hpirsiav/papers/ADLdataset/}}. The actions were not described in detail to ensure a varied dataset and each recording is around 30 minutes, resulting in a total over over 10 hours. The videos are annotated with the 18 action labels, as well as object bounding boxes (42 classes), object ids and human-object interactions. As a result, this dataset is suitable for egocentric activity recognition, forecasting and, to a lesser extent, object tracking.

\paragraph{CMU Panoptics dataset }
The CMU Panoptics dataset\footnote{\url{http://domedb.perception.cs.cmu.edu/}} \cite{joo_panoptic_2016} focuses on multi-view recordings of multi-person activities to study the 3D motion of groups engaging in social interactions. It has been collected with the aptly named ``Massively Multiview System", which contains 480 VGA camera views, 30+ HD views and 10 RGB-D sensors. In addition to the video streams, the system also provides the 3D body poses and 3D facial landmarks as well as transcripts of the performed activities including speaker ID for instances where the actors talked. This dataset occupies an interesting niche as there is no other recording setup that provides similarly detailed multi-view and 3D information.

\paragraph{CNTU RGB+D 120}
The CNTU RGB+D 120 dataset\footnote{\url{https://github.com/shahroudy/NTURGB-D}} \cite{liu_ntu_2019} is an extension of the CNTU RGB+D dataset, a large-scale dataset for RGB+D human action recognition. It contains 106 subjects performing actions for more than 114 thousand video samples and 8  million frames. Among the 120 action classes are drinking water, eating a meal, brushing teeth and hair etc. The authors introduced the dataset in CVPR 2016~\cite{shahroudy_ntu_2016} (first iteration) and later released an extended version in TPAMI 2019~\cite{liu_ntu_2019} and use it to showcase their action recognition framework.

\paragraph{JPL-Interaction dataset}
The JPL First-Person Interaction dataset (JPL-Interaction dataset)\footnote{\url{http://michaelryoo.com/jpl-interaction.html}} \cite{ryoo_first-person_2013} is geared towards researchers that are interested in egocentric social activity. It consists of human activity videos taken from  the viewpoint of one person in a multi-person group. The focus is on first-person interactive activities such as shaking hands, hugging and waving. The footage was captured by outfitting the observer (a teddy bear) with a gopro and having others interact. To remove the influence of humanoid walking patterns, the subject with the camera was placed on wheels and pushed - similarly to how a wheeled robot would move. It is one of the few datasets in this list that tackle the notoriously subjective and hard to evaluate area of social interaction.
\begin{table}[t]
\resizebox{\columnwidth}{!}{%
{\renewcommand\arraystretch{1.25}
\begin{tabular}{|l|l|l|l||l|l|l|l|l|l|l|l|} \hline
\multicolumn{2}{||p{2cm}|}{Dataset} & \multicolumn{2}{||p{2cm}|}{research purpose} &\multicolumn{2}{p{2cm}|}{data format} & \multicolumn{2}{p{2.5cm}|}{annotation format}  & \multicolumn{2}{p{2.5cm}|}{size}  & \multicolumn{2}{p{2.5cm}|}{perspective of recording}\\ \hline\hline

\multicolumn{2}{||p{2cm}|}{Kinetics \cite{kay_kinetics_2017}} & \multicolumn{2}{||p{2cm}|}{activity recognition} & \multicolumn{2}{p{2cm}|}{\raggedright videos}  & \multicolumn{2}{p{2.5cm}|}{\raggedright activity with start and stop} & \multicolumn{2}{p{2.5cm}|}{\raggedright around 1800 hours, over 420000 videos} & \multicolumn{2}{p{2.5cm}|}{\raggedright third person}\\ \hline

\multicolumn{2}{||p{2cm}|}{CMU-MMAC \cite{torre_guide_2008}} & \multicolumn{2}{||p{2cm}|}{activity and pose estimation} & \multicolumn{2}{p{2cm}|}{\raggedright videos}  & \multicolumn{2}{p{2.5cm}|}{\raggedright activity with start and stop} & \multicolumn{2}{p{2.5cm}|}{\raggedright videos of 18 actors making 5 recipes} & \multicolumn{2}{p{2.5cm}|}{\raggedright third and first person}\\ \hline

\multicolumn{2}{||p{2cm}|}{ActivityNet \cite{fabian_caba_heilbron_activitynet_2015}} & \multicolumn{2}{||p{2cm}|}{activity recognition}  & \multicolumn{2}{p{2cm}|}{\raggedright videos} & \multicolumn{2}{p{2.5cm}|}{\raggedright activity with start and stop, parent activities} & \multicolumn{2}{p{2.5cm}|}{\raggedright around 849 hours, about 28000 videos} & \multicolumn{2}{p{2.5cm}|}{\raggedright third person}\\ \hline

\multicolumn{2}{||p{2cm}|}{Charades-Ego \cite{sigurdsson_hollywood_2016}} & \multicolumn{2}{||p{2cm}|}{activity recognition}  & \multicolumn{2}{p{2cm}|}{\raggedright videos} & \multicolumn{2}{p{2.5cm}|}{\raggedright activity with start and stop} & \multicolumn{2}{p{2.5cm}|}{\raggedright 7,860 videos} & \multicolumn{2}{p{2.5cm}|}{\raggedright third and first person}\\ \hline

\multicolumn{2}{||p{2cm}|}{Cornell Activity \cite{sung_human_2011}} & \multicolumn{2}{||p{2cm}|}{activity recognition}  & \multicolumn{2}{p{2cm}|}{\raggedright videos with depth} &  \multicolumn{2}{p{2.5cm}|}{\raggedright activity segments, skeletons} & \multicolumn{2}{p{2.5cm}|}{\raggedright 180 videos together} & \multicolumn{2}{p{2.5cm}|}{\raggedright third person}\\ \hline

\multicolumn{2}{||p{2cm}|}{EPIC-Kitchens-100 \cite{damen_rescaling_2020}} & \multicolumn{2}{||p{2cm}|}{activity recognition}  & \multicolumn{2}{p{2cm}|}{\raggedright videos} & \multicolumn{2}{p{2.5cm}|}{\raggedright spoken annotations, masks and bounding boxes} & \multicolumn{2}{p{2.5cm}|}{\raggedright 100 hours of footage} & \multicolumn{2}{p{2.5cm}|}{\raggedright first person}\\ \hline

\multicolumn{2}{||p{2cm}|}{EGTEA Gaze+ \cite{li_eye_2018}} & \multicolumn{2}{||p{2cm}|}{activity recognition}  & \multicolumn{2}{p{2cm}|}{\raggedright videos} & \multicolumn{2}{p{2.5cm}|}{\raggedright activity class , hand masks and audio} & \multicolumn{2}{p{2.5cm}|}{\raggedright 28 hours of footage} & \multicolumn{2}{p{2.5cm}|}{\raggedright first person}\\ \hline

\multicolumn{2}{||p{2cm}|}{ADL Dataset \cite{pirsiavash_detecting_2012}} & \multicolumn{2}{||p{2cm}|}{activity recognition}  & \multicolumn{2}{p{2cm}|}{\raggedright videos} & \multicolumn{2}{p{2.5cm}|}{\raggedright activity class, objects with id and interaction} & \multicolumn{2}{p{2.5cm}|}{\raggedright 10+ hours of footage} & \multicolumn{2}{p{2.5cm}|}{\raggedright first person}\\ \hline

\multicolumn{2}{||p{2cm}|}{CMU Panoptics \cite{joo_panoptic_2016}} & \multicolumn{2}{||p{2cm}|}{activity and pose estimation} & \multicolumn{2}{p{2cm}|}{\raggedright videos, pose and face} &  \multicolumn{2}{p{2.5cm}|}{\raggedright annotations, speaker iD} & \multicolumn{2}{p{2.5cm}|}{\raggedright 65 sequences (5.5 hours)} & \multicolumn{2}{p{2.5cm}|}{\raggedright third person}\\ \hline

\multicolumn{2}{||p{2cm}|}{CNTU RGB+D 120 \cite{liu_ntu_2019}} & \multicolumn{2}{||p{2cm}|}{activity recognition}  & \multicolumn{2}{p{2cm}|}{\raggedright videos} & \multicolumn{2}{p{2.5cm}|}{\raggedright activity class annotations} & \multicolumn{2}{p{2.5cm}|}{\raggedright 114,480 videos, 120 classes, 106 subjects} & \multicolumn{2}{p{2.5cm}|}{\raggedright third person}\\  \hline

\multicolumn{2}{||p{2cm}|}{JPL-Interaction dataset \cite{ryoo_first-person_2013} } & \multicolumn{2}{||p{2cm}|}{activity recognition}  & \multicolumn{2}{p{2cm}|}{\raggedright videos} & \multicolumn{2}{p{2.5cm}|}{\raggedright activity class annotations} & \multicolumn{2}{p{2.5cm}|}{\raggedright 84 videos} & \multicolumn{2}{p{2.5cm}|}{\raggedright first person}\\  \hline

\end{tabular}}
}
\caption{Datasets for activity recognition.}
\label{activity_table}
\end{table}  

\subsection{Trajectory forecasting}
A lot of early work on social robot navigation among humans has focused on trajectory forecasting and achieved impressive results because even when the robot does not necessarily interact with humans, it is still perceived as socially capable as long as it does not ``get in the way". In addition, trajectory prediction is crucial for autonomous vehicles and other safety applications. \\
Alahi et al. \cite{alahi_social_2016} presented a forecasting model based on LSTMs, Kim et al. \cite{kim_predicting_2014} use Kalman filters and Maximum Likelihood estimations and Xiao et al. \cite{xiao_unsupervised_2015} use a pretrained SVM to group activity classes, predict the trajectories based on those classes and test the result in a real kitchen. Solutions that use deep learning and reinforcmeent learning have been suggested by Perez et al. \cite{perez_higueras_learning_2018}  and Fahad et al. \cite{fahad_learning_2018} respectively. The work by Pfeiffer et al. \cite{pfeiffer_data-driven_2018} takes cluttered environments into consideration and the approach by Chung et al.~\cite{chung_mobile_2010}  builds a spatial behavior cognition model to model how human trajectories influence each other. The recent work by Vemula et al.~\cite{vemula_social_2018} proposed an attention-based model that utilizes recurrent neural networks.\\
For more in depth reading we recommend the survey by Rudenko et al. \cite{rudenko_human_2020}, who have also extended the idea of human trajectory prediction to the problem of occupancy prediction \cite{Rudenko2021}: instead of predicting individual trajectories, they infer pedestrian motion from semantic information about the environment.

\subsubsection{Socially aware Navigation using Trajectory Analysis} 
Early work by Henry et al. \cite{henry_learning_2010}  uses simulated crowds to generate human-like motion behaviour while \cite{Svenstrup2010TrajectoryPF} proposed a Trajectory planning algorithm based on Rapidly-exploring Random Trees that was tested in simulation. The work by O'Callaghan~\cite{ocallaghan_learning_2011}, in contrast, does not aim to explicitly create human-like trajectories but rather to obey a set of social rules and the generated trajectories are human-like as a result. Luber et al.~\cite{luber_socially-aware_2012} learn motion prototypes from extracted ground truth behaviour and demonstrate superiority over a purely proxemics-based baseline. The work by Ferrer et al.~\cite{ferrer_robot_2013} focuses more on modeling human-agent interaction with the popular social force model, introduced by Helbing et al. \cite{helbing_social_1998}, and human feedback. A similar direction was pursued by Shiomi et al.~\cite{shiomi_towards_2014} when avoiding robot-human collisions. Tai et al.~\cite{tai_socially_2018} used GANs for behaviour cloning and Riaz et al.~\cite{riaz_collision_2018} proposed a collision avoidance system for automotive applications that is based on human behaviour. Rudenko et al. \cite{Rudenko2018} used a MDP (Markov Decision Process) algorithm with a Social Force model that emphasised the the influence of group motion over individual motion. See Table \ref{socially_aware_treajectory_table} for a comparison.\\

\begin{table}[t]
\resizebox{\columnwidth}{!}{%
{\renewcommand\arraystretch{1.25}
\begin{tabular}{|l|l|l|l||l|l|l|l|l|l|l|l|} \hline
\multicolumn{2}{||p{2cm}|}{Publication} & \multicolumn{2}{||p{2cm}|}{problem definition as written in the paper} &\multicolumn{2}{p{2cm}|}{underlying algorithm} & \multicolumn{2}{p{2.5cm}|}{used dataset}  & \multicolumn{2}{p{2.5cm}|}{used metrics}  & \multicolumn{2}{p{2.5cm}|}{tests with real humans}\\ \hline\hline

\multicolumn{2}{||p{2cm}|}{\cite{henry_learning_2010}} & \multicolumn{2}{||p{2cm}|}{generate human-like motion behavior} & \multicolumn{2}{p{2cm}|}{\raggedright simulation by \cite{Treuille2006} }  & \multicolumn{2}{p{2.5cm}|}{\raggedright inverse Reinforcement Learning (IRL)} & \multicolumn{2}{p{2.5cm}|}{\raggedright mean and max distance to ground truth from simulator} & \multicolumn{2}{p{2.5cm}|}{\raggedright no}\\ \hline

\multicolumn{2}{||p{2cm}|}{\cite{ocallaghan_learning_2011}} & \multicolumn{2}{||p{2cm}|}{navigate in a human-like manner} & \multicolumn{2}{p{2cm}|}{\raggedright Gaussian process}  & \multicolumn{2}{p{2.5cm}|}{\raggedright  UTS  RobotAssist  Project, Edinburgh Informatics Forum Pedestrian Database \cite{majecka_statistical_2009}} & \multicolumn{2}{p{2.5cm}|}{\raggedright  Cross-Validation} & \multicolumn{2}{p{2.5cm}|}{\raggedright no}\\ \hline

\multicolumn{2}{||p{2cm}|}{\cite{luber_socially-aware_2012}} & \multicolumn{2}{||p{2cm}|}{socially-aware
navigation among people} & \multicolumn{2}{p{2cm}|}{\raggedright Unsupervised Learning of Motion Prototypes}  & \multicolumn{2}{p{2.5cm}|}{\raggedright Edinburgh Informatics Forum Pedestrian Database \cite{majecka_statistical_2009}} & \multicolumn{2}{p{2.5cm}|}{\raggedright path length and similarity between the
planned path and ground truth path} & \multicolumn{2}{p{2.5cm}|}{\raggedright no}\\ \hline

\multicolumn{2}{||p{2cm}|}{\cite{ferrer_robot_2013}} & \multicolumn{2}{||p{2cm}|}{human-aware
navigation} & \multicolumn{2}{p{2cm}|}{\raggedright Social Force Model}  & \multicolumn{2}{p{2.5cm}|}{\raggedright proprietary simulation} & \multicolumn{2}{p{2.5cm}|}{\raggedright success (position goal reached)} & \multicolumn{2}{p{2.5cm}|}{\raggedright yes}\\ \hline

\multicolumn{2}{||p{2cm}|}{\cite{helbing_social_1998}} & \multicolumn{2}{||p{2cm}|}{modeling of pedestrian motion} & \multicolumn{2}{p{2cm}|}{\raggedright Social Force Model (origin paper)}  & \multicolumn{2}{p{2.5cm}|}{\raggedright proprietary simulation} & \multicolumn{2}{p{2.5cm}|}{\raggedright none (only qualitative observations)} & \multicolumn{2}{p{2.5cm}|}{\raggedright no}\\ \hline

\multicolumn{2}{||p{2cm}|}{\cite{shiomi_towards_2014}} & \multicolumn{2}{||p{2cm}|}{socially acceptable navigation behavior} & \multicolumn{2}{p{2cm}|}{\raggedright proprietary dataset collected from experiments}  & \multicolumn{2}{p{2.5cm}|}{\raggedright Social Force Model} & \multicolumn{2}{p{2.5cm}|}{\raggedright percentage of unsafe behaviour} & \multicolumn{2}{p{2.5cm}|}{\raggedright yes}\\ \hline

\multicolumn{2}{||p{2cm}|}{\cite{tai_socially_2018}} & \multicolumn{2}{||p{2cm}|}{socially compliant navigation} & \multicolumn{2}{p{2cm}|}{\raggedright collected dataset and simulation}  & \multicolumn{2}{p{2.5cm}|}{\raggedright generative adversarial imitation
learning} & \multicolumn{2}{p{2.5cm}|}{\raggedright minimal robot-pedestrian
distance and travel time from start to goal} & \multicolumn{2}{p{2.5cm}|}{\raggedright yes}\\ \hline

\multicolumn{2}{||p{2cm}|}{\cite{riaz_collision_2018}} & \multicolumn{2}{||p{2cm}|}{collision avoidance} & \multicolumn{2}{p{2cm}|}{\raggedright simulation (Netlogo) }  & \multicolumn{2}{p{2.5cm}|}{\raggedright Social Norms} & \multicolumn{2}{p{2.5cm}|}{\raggedright average number of accidents} & \multicolumn{2}{p{2.5cm}|}{\raggedright no}\\ \hline

\multicolumn{2}{||p{2cm}|}{\cite{Svenstrup2010TrajectoryPF}} & \multicolumn{2}{||p{2cm}|}{ Trajectory planning for a robot in dynamic human environments} & \multicolumn{2}{p{2cm}|}{\raggedright proprietary simulation }  & \multicolumn{2}{p{2.5cm}|}{\raggedright Rapidly-exploring Random Trees} & \multicolumn{2}{p{2.5cm}|}{\raggedright percentage of successful navigation (no collision)} & \multicolumn{2}{p{2.5cm}|}{\raggedright no}\\ \hline

\multicolumn{2}{||p{2cm}|}{\cite{Rudenko2018}} & \multicolumn{2}{||p{2cm}|}{prediction of human motion in populated space} & \multicolumn{2}{p{2cm}|}{\raggedright ATC shopping center dataset \cite{brscic_person_2013} }  & \multicolumn{2}{p{2.5cm}|}{\raggedright MDP and Social Force Model} & \multicolumn{2}{p{2.5cm}|}{\raggedright Negative Log-Probability and modified Hausdorff Distance} & \multicolumn{2}{p{2.5cm}|}{\raggedright no}\\ \hline

\end{tabular}}
}
\caption{Comparison of the Algorithms that use Trajectory Modeling for Socially aware Navigation.}
\label{socially_aware_treajectory_table}
\end{table}

\subsubsection{Datasets for Trajectory Prediction}
We will discuss a select few datasets that are used to benchmark trajectory prediction algorithms - our focus will be on the prediction of pedestrian movement, see Table \ref{trajectory_table}. Another research sector that is very interested in trajectory prediction is the automotive sector. Examples for those types of datasets that contain cars, buses, cyclists and pedestrians are Argoverse\footnote{\url{https://www.argoverse.org/}}, inD\footnote{\url{https://www.ind-dataset.com/}}, Lyft Level 5\footnote{\url{https://self-driving.lyft.com/level5/data/}} and TRAF\footnote{\url{https://gamma.umd.edu/researchdirections/autonomousdriving/trafdataset}}. The focus of this survey, as previously mentioned, is visual perception and as a result we will not go into details on datasets created with Lidar or other 3D sensors, such as the very big shopping center dataset by Brscic et al. \cite{brscic_person_2013}. \\
We recommend this github repository\footnote{\url{https://github.com/crowdbotp/OpenTraj}} for an overview that includes traffic and non-traffic setups as well as the TrajNet challenge/benchmark\footnote{\url{http://trajnet.stanford.edu/}} \cite{sadeghian_trajnet_2018}, that combines a number of the mentioned datasets and offers tools for evaluations across datasets. \\
Something to keep in mind with these datasets is that the authors rarely obtain the informed consent of the people in the videos and as a result they might not be available in the future (see for example the Oxford Town Centre dataset\footnote{\url{https://megapixels.cc/oxford_town_centre/}}), and that in most cases the perspective from which the data is collected is not a potential agent's egocentric view (most datasets are captured from a very high vantage point).

\paragraph{ETH dataset}

The ETH BIWI Walking Pedestrians (EWAP)\footnote{\url{https://icu.ee.ethz.ch/research/datsets.html}} dataset is comprised of videos showing pedestrians in unconstrained situations, shot from a high vantage point. It was recorded at 25 fps and manually annotated at 2.5 fps, totalling 650 human trajectories over 25 minutes. The dataset contains videos from two different vantage points and was used for the ICCV 2009 paper \cite{pellegrini_youll_2009} that presents a solution to the task of multi-person tracking in crowded environments.

\paragraph{Crowds UCY dataset}

This dataset contains aerial view videos of everyday crowd scenarios. The videos show students, shoppers and pedestrians in general and are shot at 2.5 fps. The dataset shows the trajectories of roughly 200 pedestrians and was presented by Lerner et al. in 2007 \cite{lerner_crowds_2007}. Although this dataset is not very new, it remains one of the more popular ones and is often used in conjunction with the ETH dataset.

\paragraph{Train Station Dataset}

The Train Station dataset or Grand Central Station dataset was presented by Zhou et al. in \cite{zhou_understanding_2012} and contains, as the name suggests, recordings of pedestrians moving freely at the Grand Central Station in New York. The surveillance video was taken from above and shows over 33 minutes at 25 fps, at a resolution of 720x480. It contains keypoint trajectories for each of the recorded pedestrians. Although it is small, the specific environment and the number of pedestrians make it interesting for research on areas with high foot traffic.

\paragraph{Edinburgh Informatics Forum Pedestrian Database}

This dataset\footnote{\url{http://homepages.inf.ed.ac.uk/rbf/FORUMTRACKING/}} consists of ``compressed" trajectories, observed by a top down camera at the University of Edinburgh over a timespan of several months. The dataset does not contain video frames like the other datasets, instead the trajectories are described by the position of the bounding box marking the detection and a histogram representation of the detected pedestrian. Given the large timeframe, the authors \cite{majecka_statistical_2009} managed to extract more than 7 million pedestrian detections and as a result the dataset contains more than 90.000 trajectories. The viewpoint of the dataset is particularly interesting as it covers not only the free interaction between pedestrians but also the behaviour at building entrances.

\paragraph{Stanford Drone Dataset }

The aim of this dataset\footnote{\url{https://cvgl.stanford.edu/projects/uav_data/}} is to showcase the interaction of pedestrians, skateboarders, cars, buses, and golf carts in various environments on the Stanford Campus. The authors \cite{robicquet_learning_2016} recorded 8 different scenes from above to explore how social rules and common sense influences human trajectories. It contains 60 videos with annotations for the mentioned classes and is currently one of the bigger datasets that are available for trajectory forecasting.

\begin{table}[t]
\resizebox{\columnwidth}{!}{%
{\renewcommand\arraystretch{1.25}
\begin{tabular}{|l|l|l|l||l|l|l|l|l|l|l|l|} \hline
\multicolumn{2}{||p{2cm}|}{Dataset} & \multicolumn{2}{||p{2cm}|}{research purpose} &\multicolumn{2}{p{2cm}|}{annotations} & \multicolumn{2}{p{2.5cm}|}{participants}  & \multicolumn{2}{p{2.5cm}|}{size}  & \multicolumn{2}{p{2.5cm}|}{perspective of recording}\\ \hline\hline

\multicolumn{2}{||p{2cm}|}{ETH Pedestrian Dataset \cite{pellegrini_youll_2009} } & \multicolumn{2}{||p{2cm}|}{multi-person tracking in crowded environments} & \multicolumn{2}{p{2cm}|}{\raggedright pedestrian IDs}  & \multicolumn{2}{p{2.5cm}|}{\raggedright pedestrians} & \multicolumn{2}{p{2.5cm}|}{\raggedright 650 trajectories over 25 minutes} & \multicolumn{2}{p{2.5cm}|}{\raggedright from above}\\ \hline

\multicolumn{2}{||p{2cm}|}{Crowds UCY Dataset \cite{lerner_crowds_2007}} & \multicolumn{2}{||p{2cm}|}{crowd simulation} & \multicolumn{2}{p{2cm}|}{\raggedright pedestrian IDs}  & \multicolumn{2}{p{2.5cm}|}{\raggedright pedestrians} & \multicolumn{2}{p{2.5cm}|}{\raggedright 200 pedestrian trajectories} & \multicolumn{2}{p{2.5cm}|}{\raggedright from above}\\ \hline

\multicolumn{2}{||p{2cm}|}{Train Station Dataset \cite{zhou_understanding_2012} } & \multicolumn{2}{||p{2cm}|}{crowd behaviour analysis} & \multicolumn{2}{p{2cm}|}{\raggedright key-point trajectories}  & \multicolumn{2}{p{2.5cm}|}{\raggedright commuters} & \multicolumn{2}{p{2.5cm}|}{\raggedright 33 minutes of footage} & \multicolumn{2}{p{2.5cm}|}{\raggedright from above}\\ \hline

\multicolumn{2}{||p{2cm}|}{Edinburgh Informatics Forum Pedestrian Dataset \cite{majecka_statistical_2009}} & \multicolumn{2}{||p{2cm}|}{pedestrian behaviour in open space and around entries} & \multicolumn{2}{p{2cm}|}{\raggedright compressed trajectories}  & \multicolumn{2}{p{2.5cm}|}{\raggedright pedestrians} & \multicolumn{2}{p{2.5cm}|}{\raggedright over 90.000 trajectories} & \multicolumn{2}{p{2.5cm}|}{\raggedright from above}\\ \hline

\multicolumn{2}{||p{2cm}|}{Stanford Drone Dataset \cite{robicquet_learning_2016}} & \multicolumn{2}{||p{2cm}|}{interaction of various traffic participants} & \multicolumn{2}{p{2cm}|}{\raggedright bounding boxes on the videos}  & \multicolumn{2}{p{2.5cm}|}{\raggedright traffic participants (pedestrians, skateboarders, cars etc.)} & \multicolumn{2}{p{2.5cm}|}{\raggedright 60 videos with 930.000 annotated frames and 20.000 detected traffic participants} & \multicolumn{2}{p{2.5cm}|}{\raggedright from above}\\ \hline

\multicolumn{2}{||p{2cm}|}{PETS 2009 Dataset \cite{ferryman_pets2010_2010}} & \multicolumn{2}{||p{2cm}|}{pedestrian detection, trajectory estimation, crowd event detection} & \multicolumn{2}{p{2cm}|}{\raggedright depending on the subset: detection of pedestrians, crowd density and person count}  & \multicolumn{2}{p{2.5cm}|}{\raggedright pedestrians} & \multicolumn{2}{p{2.5cm}|}{\raggedright 5 GB of 768x576 RGB video} & \multicolumn{2}{p{2.5cm}|}{\raggedright from above}\\ \hline

\end{tabular}}
}
\caption{ Datasets for trajectory forecasting.}
\label{trajectory_table}
\end{table}

\paragraph{PETS 2009 Dataset}

This dataset\footnote{\url{http://cs.binghamton.edu/~mrldata/pets2009}} was presented by Ferryman et al. \cite{ferryman_pets2010_2010} in 2010 as part of a workshop on pedestrian detection, counting and trajectory estimation as well as crowd event detection. Given the different tasks that the dataset is concerned with, it is divided into subsets for specific problem settings. One subset is meant for general training, one is for benchmarking person count and density estimation, another is aimed at pedestrian tracking and the third is for evaluating event recognition and flow analysis algorithms. The dataset consists of RGB videos of students and pedestrians moving across the Whiteknights Campus, University of Reading, taken from an elevated view.

\section{Discussion}
\begin{figure}
\resizebox{\columnwidth}{!}{%
\includegraphics[width=\textwidth]{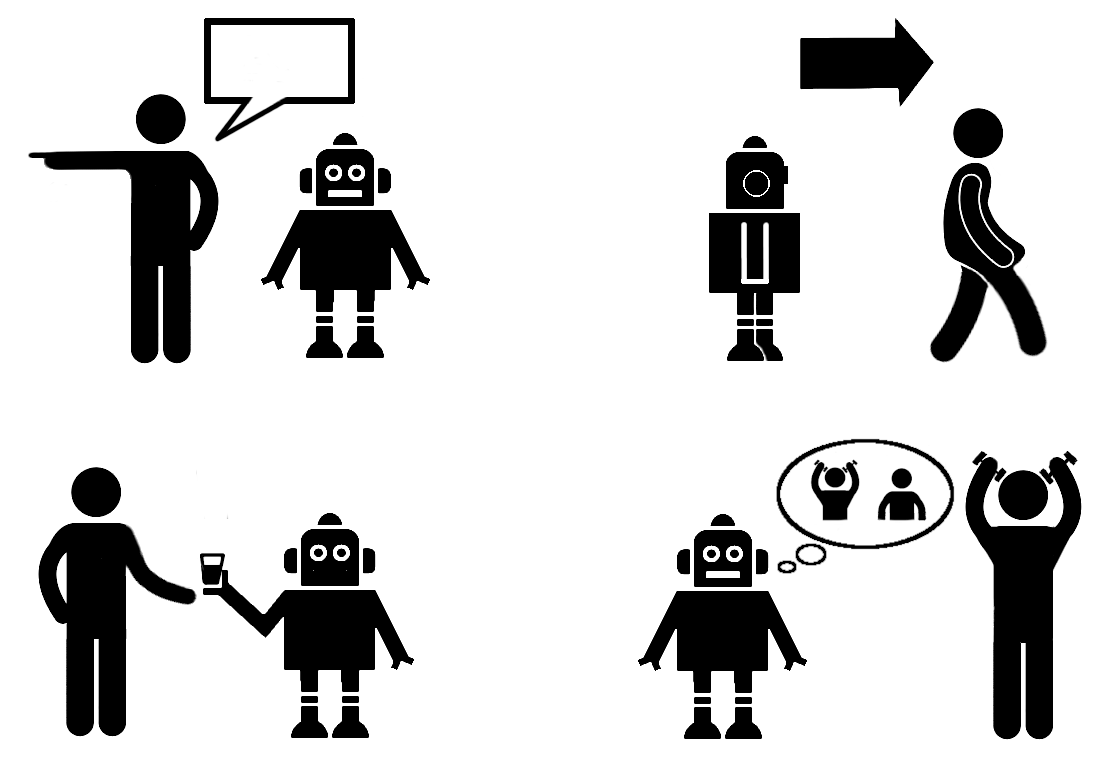}
}
\caption{ Illustration of potential benchmarking tasks for Socially ware Robot Navigation - Embodied Question Answering with a human asking the question (upper left), Human Following (upper right), handover tasks (lower left) and continuous observation (lower right). }
\label{social_subtasks}
\end{figure}

We will highlight observed trends or research gaps in this section. In many cases, the underlying algorithms or feature representations already exist, but they are not yet widely used to enrich algorithms for Socially Aware Robot Navigation.
\paragraph{Formalised Evaluation Strategies}
Unlike goal-oriented Robot Navigation or Trajectory Prediction, Socially aware Robot Navigation is difficult to evaluate in terms of objective metrics. Not only does the robot have to perform a certain task, but the positive or negative effect it has on the humans sharing the space has to be taken into account as well. Often this is done with small scale human studies, where the users are asked to rate the behaviour in terms of subjective qualities such as naturalness or comfort \cite{charalampous_recent_2017}. Without formalised metrics for social behaviour, different algorithms are hard to compare but the implementation of such a metric is borderline impossible due to the inherently subjective nature of the task. An alternative might be not to evaluate if a robot behaves overall socially compliant, but whether it can perform basic subtasks that are required in daily interactions - just like the clearly defined subtasks of PointGoal Navigation and Coverage based Exploration are used to gauge whether a robot can navigate intelligently. \\
Possible subtasks for Socially aware robot Navigation might be Human Following \cite{Xiao2020} \cite{tee_kit_tsun_exploring_2018}, handover tasks \cite{Strabala2013}, continuous observation or Embodied Question asking with spoken instructions from a human, see Figure \ref{social_subtasks}\\
The task of human following in particular is a very specific and self-contained subset of the overall topic but it is one of the most useful and comparatively easy to evaluate. Especially in settings where robots are used as transport vehicles to unburden the human user, it might be beneficial to implement a following strategy. \\
Petrovic et al. \cite{petrovic_stereo_2013} present a solution based on stereo vision and Kalman filters that is capable of following the user in indoor and outdoor scenarios. A solution with cameras in addition to laser range sensors was presented by Shimizu et al.~\cite{shimizu_mobile_2012}, who achieve markerless results in indoor settings. In contrast, Kit Tsun et al. \cite{tee_kit_tsun_exploring_2018} use a depth camera, an active marker and proximity sensors as well as simulated environments. Two approaches using reinforcement learning have been proposed by Wang et al. \cite{wang_reinforcement_2006} and Bayoumi et al. \cite{bayoumi_efficient_2015}, with the former learning the correct social distance when interacting with the human and the latter learning about the human's goal so the robot can navigate efficiently even when the human makes detours. \\
A more formal definition and shared understanding of the Human Following task (for example: defined distance to the human, defined input modalities) and corresponding metrics (for example: percentage of steps with the human in ``frame" for visual input, percentage of steps at the correct distance, penalty for invading the person´s space) could elevate this task to a shared benchmark.

\paragraph{Interaction via Language}
Language is already used for general Human Robot Interaction \cite{khayrallah2015natural}, but it could also be used to convey instructions or additional information for Socially Aware Robot Navigation. Recent research on Robot Navigation has seen an uptick in language-guided Navigation \cite{Zhou2020},\cite{Chi_Shen_Eric_Kim_Hakkani-tur_2020}, since it allows for complex semantic task descriptions. The focus of those works are usually on the conversion of the written language task into a concrete goal formulation for the robot and this scope could be easily extended to a version where the human is in the loop. This would also entail strategies for good information transfer (for example: good listening positions, conveying attention). Similarly, Embodied Question Answering \cite{das_embodied_2017}, another well-defined subtask of Robot Navigation, could be adapted to have a human pose the question verbally. 

\paragraph{Gesture and Emotion Recognition}
Gesture Recognition is used in the context of HRI \cite{Xu2014} but gestures are also a natural way to enrich navigation behaviours \cite{Lei2014}, for example the nonverbal communication of ``after you" at bottlenecks or ``come here" gestures. An adjacent topic is the integration of emotional cues into Human Robot Interaction \cite{correa_face_2008} and potentially robot navigation. Narayanan et al. \cite{narayanan_proxemo_2020} inferred overall mood from gaits, which were used to adapt the robot´s path and Bera et al. \cite{Bera2020} used trajectories and faces, as did Gines et al. \cite{gines2019social}. Yelwande et al. \cite{Yelwande2020} compared different path planning strategies that utilise information about the human´s emotional state.\\
\paragraph{Datasets}
Socially Aware Robot navigation is a difficult topic to create realistic datasets for, since they have to encompass the rich input of an embodied agent, the variations of human behaviour and the fact that the behaviour of the robot itself will affect the humans around it. In addition, one dataset will not necessarily fit robots that are different in size and shape, since this also influences how humans react, which makes recording reliable training data a difficult task. Simulated datasets on the other hand are widely used, but they will always suffer bias in terms of the representation of the world as well as the human behaviour. Nevertheless, we have seen the first promising datasets that are explicitly created to model human-aware robot navigation such as the dataset by Tolani et al. \cite{tolani_visual_2020} or the very recent SEAN environment \cite{tsoi_sean_2020}\footnote{\url{https://sean.interactive-machines.com/setup.html}}. However both are still very new and not widely used yet. Another recent work by Baghel et al.~\cite{baghel_toolkit_2020} also proposes the design of simulation frameworks specifically for social navigation, although it focuses more on trajectory simulation than realistic visual input. Similarly, Puig et al.~\cite{puig_virtualhome_2018} presented a framework for modeling human activities with the aim of simulating daily task and potentially training collaborative behaviour \cite{puig_watch-and-help_2020}. All of this is indicative of the desire to pursue this area of research, even when bigger and more influential frameworks such as AI Habitat and AI2 Thor do not support human actors yet. However, a recent announcement from the team of the iGibson simulator suggests that Socially aware Navigation will be of interest to them: starting with the iGibson Challenge 2021, the robot can be trained to navigate in simulated crowds \footnote{http://svl.stanford.edu/igibson/challenge.html}.\\

\noindent
In summary, we have covered motivations for socially-aware robot navigation, ranging from industrial contexts, such as assembly and packing, to healthcare and social companionship. A brief introduction of the Active Vision paradigm was given, since it is the basis for most of the highlighted research and for Robot Navigation in general. We also looked into the current and previous trends for Robot Navigation that are most relevant for socially-aware applications. In addition, the topics Human Behaviour Modeling and Analysis and Human Robot Interaction were briefly introduced, again with a focus on works that are relevant for our topic. The last section on  Human-aware Motion Planning in particular has highlighted how different areas of research can come together naturally to model behaviour that is suitable for a human environment. \\
We have found a variety of promising approaches and datasets that aim to solve the complex problem of socially-aware Robot Navigation, but we have also observed that at the current point in time, there are few works that focus specifically on this challenging topic. Especially with the current popularity of simulation-based learning, a variety of datasets and frameworks exist that could in theory be augmented to include humans. Instead, most approaches are confined to only Robot Navigation without humans, Human Robot Interaction without robot movement etc..\\
We hope that the overview given in this survey will serve as inspiration to other researchers for more interdisciplinary research and we look forward to future work in this topic.
\section{Acknowledgements}
This research has been supported by Marie Skłodowska-Curie Innovative Training Networks - European Industrial Doctorates - PhilHumans Project (Contract no. 812882, Web: \url{http://www.philhumans.eu}), MIUR AIM - Attrazione e Mobilita Internazionale Linea 1 - AIM1893589 - CUP: E64118002540007 and by MISE - PON I\&C 2014-2020 -  ENIGMA Project - Prog n. F/190050/02/X44 – CUP: B61B19000520008.

\balance
\bibliography{main}

\end{document}